\documentclass[iicol,pdflatex,sn-mathphys-num]{sn-jnl}

\usepackage{graphicx}%
\usepackage{multirow}%
\usepackage{amsmath,amssymb,amsfonts}%
\usepackage{amsthm}%
\usepackage{mathrsfs}%
\usepackage[title]{appendix}%
\usepackage{xcolor}%
\usepackage{textcomp}%
\usepackage{manyfoot}%
\usepackage{booktabs}%
\usepackage{algorithm}%
\usepackage{algorithmicx}%
\usepackage{algpseudocode}%
\usepackage{listings}%

\usepackage{acro}
\usepackage{array}
\usepackage{natbib}
\usepackage{placeins}

\acsetup{make-links = true, link-only-first = true}

\DeclareAcronym{eg}{short={e.g.,},long=for example,foreign=exempli gratia,foreign-babel=latin,first-style=short}
\DeclareAcronym{cpu}{short=CPU,long=Central Processing Unit,first-style=short-long,subsequent-style=short,long-plural=s,short-plural=s}
\DeclareAcronym{gpu}{short=GPU,long=Graphics Processing Unit,first-style=short-long,subsequent-style=short,long-plural=s,short-plural=s}
\DeclareAcronym{png}{short=PNG,long=Portable Network Graphics,first-style=short-long,subsequent-style=short}
\DeclareAcronym{ai}{short=AI,long=Artificial Intelligence,first-style=long-short,subsequent-style=short,long-plural=s,short-plural=s}
\DeclareAcronym{cli}{short=CLI,long=Command-line Interface,first-style=long-short,subsequent-style=short}
\DeclareAcronym{api}{short=API,long=Application Programming Interface,first-style=long-short,subsequent-style=short}
\DeclareAcronym{pdf}{short=PDF,long=Portable Document File,first-style=short-long,subsequent-style=short}
\DeclareAcronym{jpeg}{short=JPEG,long=Joint Photographic Experts Group,first-style=short-long,subsequent-style=short}
\DeclareAcronym{tiff}{short=TIFF,long=Tagged Image File Format,first-style=short-long,subsequent-style=short}
\DeclareAcronym{csv}{short=CSV,long=Comma-separated Values,first-style=short-long,subsequent-style=short}
\DeclareAcronym{lda}{short=LDA,long=Latent Dirichlet Analysis,first-style=long-short,subsequent-style=short}
\DeclareAcronym{knn}{short=k-NN,long=k-nearest neighbors,first-style=long-short,subsequent-style=short}
\DeclareAcronym{svm}{short=SVM,long=Support Vector Machine,first-style=long-short,subsequent-style=short,long-plural=s,short-plural=s}
\DeclareAcronym{cuda}{short=CUDA,long=Compute Unified Device Architecture,first-style=long-short,subsequent-style=short}
\DeclareAcronym{dla}{short=DLA,long=Document Layout Analysis,first-style=long-short,subsequent-style=short}
\DeclareAcronym{ufal}{short=IFAL,long=Institute of Formal and Applied Linguistics,foreign=Ústav formální a aplikované lingvistiky (ÚFAL),foreign-babel=czech,first-style=long-short,subsequent-style=short}
\DeclareAcronym{arup}{short=IAP,long=Institute of Archaeology of the Czech Academy of Sciences in Prague,foreign=Archeologický ústav AV ČR Praha v. v. i. (ARÚP),foreign-babel=czech,first-style=long-short,subsequent-style=short}
\DeclareAcronym{arub}{short=IAB,long=Institute of Archaeology of the Czech Academy of Sciences in Brno,foreign=Archeologický ústav AV ČR Brno v. v. i. (ARÚB),foreign-babel=czech,first-style=long-short,subsequent-style=short}
\DeclareAcronym{dd}{short=DD,long=DeepDoctection,first-style=long-short,subsequent-style=short}
\DeclareAcronym{vit}{short=ViT,long=Vision Transformer,first-style=long-short,subsequent-style=short,long-plural=s,short-plural=s}
\DeclareAcronym{dit}{short=DiT,long=Document Image Transformer,first-style=long-short,subsequent-style=short,long-plural=s,short-plural=s}
\DeclareAcronym{cnn}{short=CNN,long=Convolutional Neural Network,first-style=long-short,subsequent-style=short,long-plural=s,short-plural=s}
\DeclareAcronym{clip}{short=CLIP,long=Contrastive Language-Image Pretraining,first-style=long-short,subsequent-style=short}
\DeclareAcronym{ocr}{short=OCR,long=Optical Character Recognition,first-style=long-short,subsequent-style=short}
\DeclareAcronym{rfc}{short=RFC,long=Random Forest Classifier,first-style=long-short,subsequent-style=short}
\DeclareAcronym{llm}{short=LLM,long=Large Language Model,long-plural=s,short-plural=s,first-style=long-short,subsequent-style=short}

\theoremstyle{thmstyleone}%
\theoremstyle{thmstyletwo}%
\theoremstyle{thmstylethree}%

\begin{document}

\title{Page image classifier fine-tuned on century-spanning archives of scanned documents for further content-specific processing}

\author*[1]{\fnm{Kateryna} \sur{Lutsai} \orcid{0009-0002-4773-2797}}\email{lutsai@ufal.mff.cuni.cz}

\author[2]{\fnm{Dana} \sur{Křivánková} \orcid{0000-0001-5718-9447}}\email{krivankova@arup.cas.cz}

\author[3]{\fnm{David} \sur{Novák} \orcid{0009-0005-8722-0245}} \email{novak@arup.cas.cz}
\equalcont{These authors contributed equally to this work.}

\author[4]{\fnm{Pavel} \sur{Straňák} \orcid{0000-0002-6895-8536}}\email{stranak@ufal.mff.cuni.cz}
\equalcont{These authors contributed equally to this work.}

\affil[1,2]{\orgdiv{Institute of Formal and Applied Linguistics}, \orgname{Charles University MFF}, \orgaddress{\street{Malostranské náměstí}, \city{Prague}, \postcode{11800}, \country{Czech Republic}}}

\affil[3]{\orgdiv{Institute of Archaeology}, \orgname{Czech Academy of Sciences}, \orgaddress{\street{Letenská}, \city{Prague}, \postcode{11800}, \country{Czech Republic}}}

\abstract{
\textbf{Purpose:} Digitization projects in the humanities produce vast, heterogeneous archives of historical documents, making manual sorting impractical at scale. This work addresses the need for an automated system to classify scanned page images based on visual content type — text, tables, and graphics — enabling content-specific downstream processing such as Optical Character Recognition (OCR) or structured data extraction.

\textbf{Methods:} An image classification system was developed and evaluated on a dataset of over 48,000 annotated historical page images from century-old Czech archaeological archives, refined through four successive annotation stages with domain-expert review. A Random Forest Classifier baseline was established using hand-crafted image features. Subsequently, deep learning architectures were fine-tuned and compared: Convolutional Neural Networks (EfficientNetV2, RegNetY), Vision and Document Image Transformers (ViT, DiT), and multimodal CLIP models. An 11-category label scheme was designed collaboratively with domain experts and evaluated via five-fold cross-validation.

\textbf{Results:} The feature-based baseline achieved approximately 75\% accuracy. Fine-tuned CNNs and Transformers substantially outperformed it, with RegNetY-16GF achieving 99.16\% and ViT-large 99.12\% Top-1 accuracy on the held-out test set. CLIP ViT-B/16 reached 99.14\% with optimized text descriptions.

\textbf{Conclusion:} Image-only models, particularly RegNetY-16GF, deliver near-perfect classification accuracy and produce consistent labels across 649,508 unlabeled archival pages with over 90\% inter-model agreement. Fine-tuned CLIP, despite competitive test-set accuracy, showed under 65\% agreement with image-only models on unlabeled data, making it less suitable for deployment. The final models, annotated dataset, and software are publicly available under open-source licenses.
}


\maketitle

\section{Introduction}\label{sec1}

\noindent The spread of large-scale digitization initiatives in libraries, archives, and museums has generated massive digital collections of historical documents. For a long time, paper materials and digitization workflows were the foundation of both newly created and old document collections, until archived documents became digital-born. While digitization of paper materials significantly improves accessibility, the management and analysis of these vast, complex repositories presents considerable challenges.

\begin{figure*}[htbp]
    \centering
    \caption{Number of page scans over time (the overall collection of unlabeled source files)}
    \label{fig:timeline}
    \includegraphics[width=\textwidth]{images/Fig1.pdf}
\end{figure*}

Digital archives derived from historical documents exhibit several unique characteristics that complicate their management. The collections often span significant historical periods, with document creation ranging from the early 20th century to the present day, and the data volume typically increases exponentially over time, as illustrated in Fig.~\ref{fig:timeline}. The archive studied here spans roughly a century (approximately 1920--2020); the ``century-old'' framing in the title refers to its earliest material, while the bulk of the collection is more recent, as Fig.~\ref{fig:timeline} makes clear. High-resolution scanning results in large image files, and the physical condition of original documents can introduce visual defects into scans.

\begin{figure*}[htbp]\centering
\includegraphics[width=130mm]{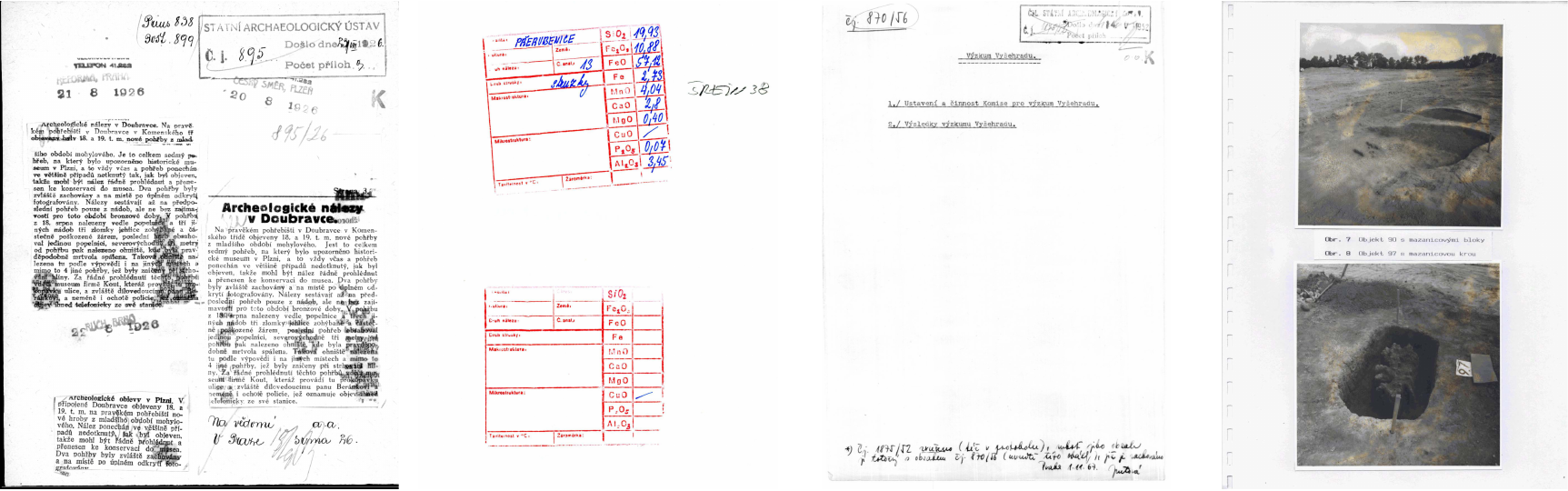}
\caption{Samples of various page types from the archive}
\label{fig:intro-page-1}
\end{figure*}

\begin{figure*}[htbp]\centering
\includegraphics[width=130mm]{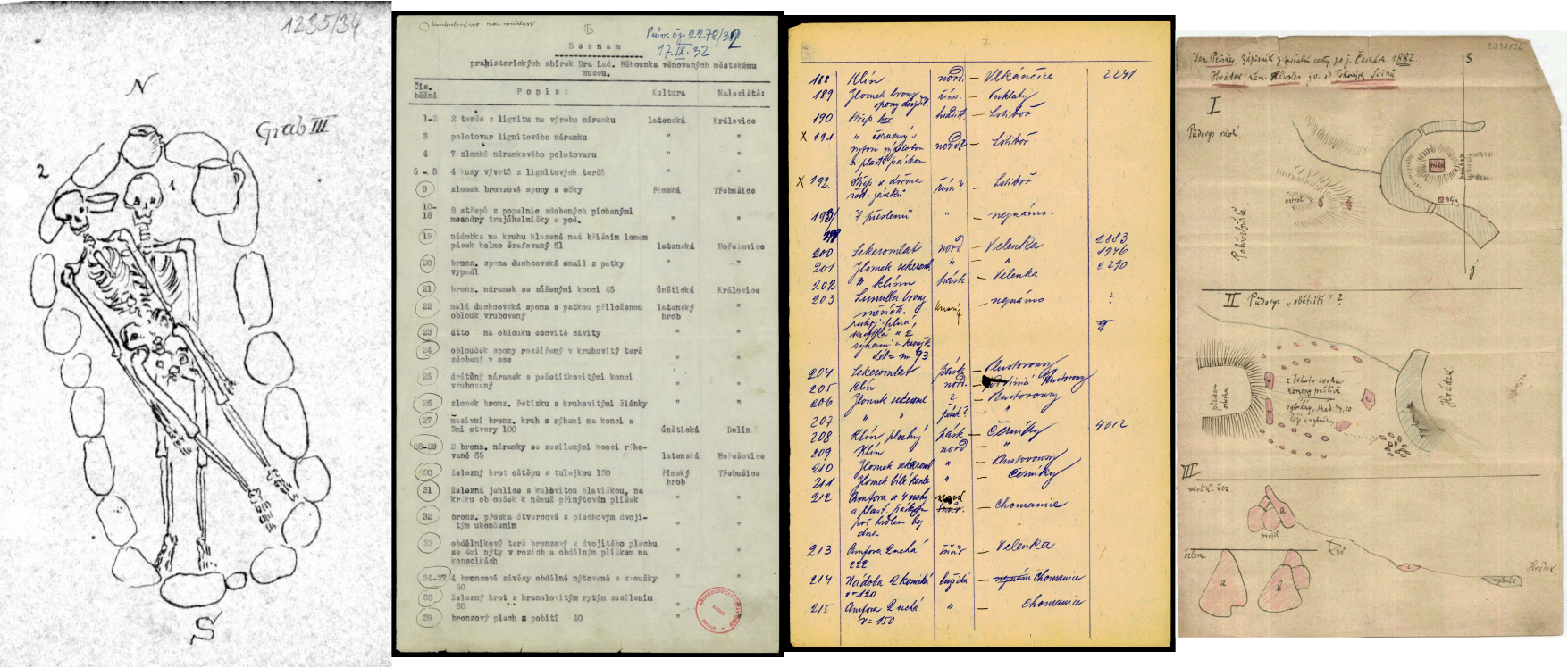}
\caption{Samples of various page types from the same collection, showing diversity in size, content, and condition}
\label{fig:intro-page-2}
\end{figure*}

Furthermore, these archives are marked by profound heterogeneity. A single collection can contain a wide variety of content types, often combined within a single document. As demonstrated in Figs.~\ref{fig:intro-page-1} and~\ref{fig:intro-page-2}, this includes everything from handwritten manuscripts and typed correspondence to printed articles, technical drawings, maps, and photographs.

In addition, scanned documents frequently suffer from metadata scarcity. File names often encode technical details, such as the scan date, rather than semantic content. Critical information like author, title, or language is often missing, complicating automated processing. This combination of diversity and poor documentation is a well-known characteristic of large-scale digitization efforts~\cite{nikolaidou2022survey}.

The characteristics outlined above create significant hurdles for effective archive management:
\begin{description}
    \item \textbf{Sorting and Organization:} The sheer volume and heterogeneity make manual sorting impractical. Scanning campaigns often generate large batches, raising the risk of human error. Processing documents with default equipment settings leads to absent descriptive metadata (language, document type, collection), preventing straightforward automated grouping.
    \item \textbf{Manual Processing Time:} Manually reviewing each page to determine its content category is prohibitively time-consuming and unsustainable at scale, a widely recognized challenge in digital humanities~\cite{nikolaidou2022survey}.
    \item \textbf{Need for Content-Specific Processing:} Different page types require different tools; \ac{ocr} is appropriate for text, layout analysis for tables~\cite{zhong2019publaynet,xu2020layoutlm}, and image segmentation for photographs. Without an initial classification step, these downstream pipelines cannot be applied efficiently.
\end{description}

Initial consultations with the data providers confirmed that the collection was perceived as highly disorganized, reflecting a common reality where the scale of data acquisition outstrips the resources available for curation.

To address these challenges, this research develops an automated system for classifying page images from historical archives based on their visual content and layout. The primary contribution is the implementation and rigorous evaluation of multiple classification approaches — from classical feature-based methods to state-of-the-art deep learning networks — on a unique, large-scale real-world archival collection. A detailed description of all experiments and implementation details is available in the accompanying thesis~\cite{lutsai2026pageimageclassification}.

\noindent\textit{LLM use disclosure.} This manuscript was prepared with assistance from large language models (Gemini 2.5, GPT-4) for text drafting and generation from author-supplied bullet-point outlines, as well as for copy editing (style, grammar, and formatting). All generated content was reviewed, verified, and revised by the authors, who take full accountability for the final text. Full details are provided in the disclosure at the end of this article.

The remainder of this article is organized as follows. Section~\ref{sec-related} reviews prior work. Section~\ref{sec2} provides a detailed overview of the raw data and its properties. Section~\ref{sec3} describes the classification methodology. In Sect.~\ref{subsec33}, the proposed system architecture is detailed. Section~\ref{sec4} presents performance results and practical integration. Section~\ref{subsec43} analyses prediction agreement on the full unlabeled collection. Finally, Sect.~\ref{sec5} concludes the paper.

\section{Background and Related Work}\label{sec-related}

\subsection{Document Image Classification}

Document image classification has been studied for several decades. Early work established standard benchmarks and demonstrated the utility of convolutional architectures. Lewis et al.~\cite{lewis2006building} constructed the IIT-CDIP collection of 42~million scanned tobacco-industry documents, which became the standard pre-training corpus for document-specific models. The RVL-CDIP benchmark~\cite{harley2015icdar}, derived from IIT-CDIP, contains 400,000 grayscale document images across 16 categories (letters, memos, forms, invoices, etc.) and has since served as the primary evaluation standard for document image classification. Initial approaches relied on hand-crafted features and classical classifiers; the survey by Liu et al.~\cite{liu2021document} traces the progression from these methods to deep convolutional networks and documents the steady accuracy gains over two decades of work.

A comprehensive survey by Nikolaidou et al.~\cite{nikolaidou2022survey} covering historical document image datasets reports that, despite the richness of available corpora, most datasets originate from well-preserved, printed Western European sources and focus on text recognition rather than holistic page-level classification. The challenge of managing genuinely heterogeneous archives — spanning handwriting, photographs, maps, and forms across more than a century — is largely unaddressed by existing benchmarks.

\subsection{Deep Learning Architectures for Document Analysis}

Convolutional neural networks have proven highly effective for document image classification. Harley et al.~\cite{harley2015icdar} showed that CNNs substantially outperform hand-crafted features on RVL-CDIP, establishing the template for subsequent fine-tuning approaches. The EfficientNetV2 family~\cite{tan2021efficientnetv2,tan2019efficientnet} and RegNet design spaces~\cite{radosavovic2020designing} offer accuracy-efficiency trade-offs well suited to large-scale batch processing: smaller variants run on standard CPUs, while larger ones match or exceed Transformer accuracy at lower computational cost.

Vision Transformers (\acf{vit}~\cite{dosovitskiy2020image,beyer2022better}) and their distillation variants~\cite{touvron2021deit} have achieved top-tier results on ImageNet~\cite{ridnik2021imagenet} and transfer competitively to document tasks when pre-trained on large corpora. The \acf{dit}~\cite{li2022dit} extended this paradigm by pre-training directly on IIT-CDIP via BEiT-style masked image modelling, reaching state-of-the-art performance on RVL-CDIP without any convolutional inductive biases.

Document layout analysis and structure understanding received further impetus from multimodal models. LayoutLM~\cite{xu2020layoutlm} jointly models textual content and spatial layout from OCR-extracted tokens, achieving strong results on form understanding and information extraction benchmarks. However, LayoutLM and its successors require reliable OCR output — a prerequisite that fails on the noisy, handwritten, and structurally degraded pages common in historical archives.

\subsection{Multimodal and Zero-Shot Classification}

\acf{clip}~\cite{radford2021learning}, pre-trained by aligning images and their natural-language captions on 400~million web image–text pairs~\cite{xu2023demystifying}, enables zero-shot image classification by scoring similarity between an image and textual category descriptions. This property is attractive for document classification, where categories can be described in natural language without requiring labeled images. However, zero-shot \ac{clip} is known to underperform on domain-specific or fine-grained visual tasks, where the pre-training distribution diverges from the target~\cite{liu2021document}. Fine-tuning recovers much of this gap, as we demonstrate in Section~\ref{subsubsec323}.

\subsection{Historical Document Analysis}

Historical document archives present additional challenges beyond those in standard benchmarks: physical degradation (yellowing, stains, tears), mixed content within a single page, metadata scarcity, and extreme temporal heterogeneity spanning a century of visual styles~\cite{nikolaidou2022survey}. Prior work on historical document processing has focused predominantly on handwriting recognition and text line segmentation rather than holistic page-level classification~\cite{nikolaidou2022survey}. The present work fills this gap by developing a domain-specific label taxonomy and a large annotated dataset targeting the heterogeneous page-level characteristics of a real-world archaeological archive.

\section{Exploration of the Raw Data} \label{sec2}

\noindent The primary dataset consists of scanned pages from the \ac{arup} (primary source) and \ac{arub} (secondary source), initially provided as multi-page \ac{pdf} documents. These were converted into individual \ac{png} image files, organized into directories corresponding to source documents. The initial data processing and exploration revealed several key characteristics of the collection.

\begin{figure*}[htbp]\centering
\includegraphics[width=130mm]{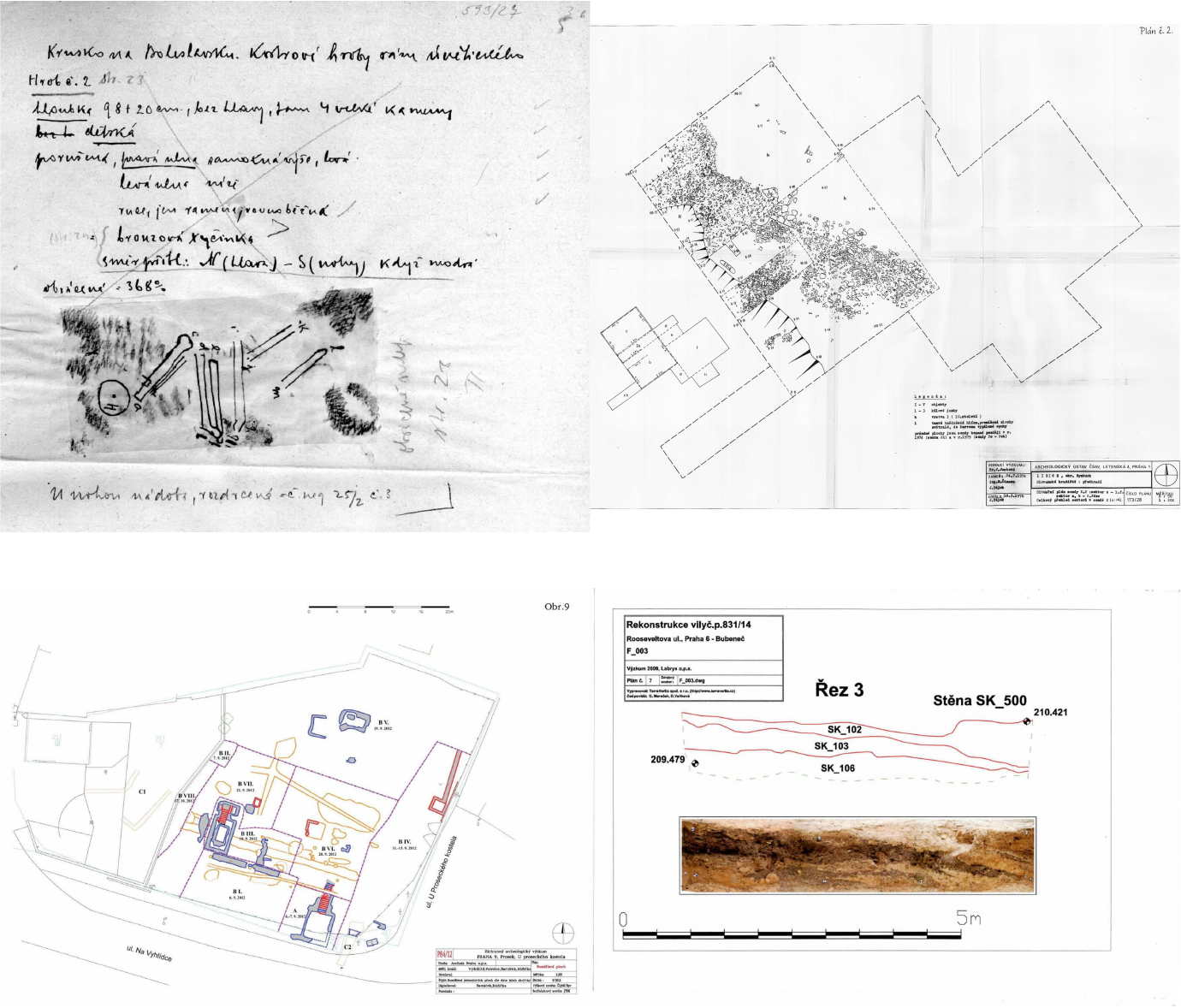}
\caption{Samples of pages from different time periods; the newer page on the right is digital-born, yet was scanned for archiving}
\label{fig:intro-time}
\end{figure*}

The archive totals approximately 400--420~GB across more than 60,000 \acp{pdf} and nearly 650,000 pages, with document creation dates spanning roughly 1920--2020. A parallel processing workflow was established on a computing cluster to convert the \ac{pdf} files into page-specific images. Large-format scans occasionally led to truncated images; these were re-processed with greater memory allocation. Following conversion, single-page documents were consolidated into a common folder, and original \ac{pdf} files were removed to conserve storage. This image-based dataset then formed the basis for exploratory analysis and manual annotation.

\subsection{Characteristics of the Source Data} \label{subsec21}

\subsubsection{Visual Defects of the Scanned Pages} \label{subsubsec211}

\noindent The scanned pages frequently exhibited visual defects originating from the physical condition of source documents and the scanning process itself. These artifacts range from minor blemishes to severe degradation that complicates automated content extraction. Representative examples of each defect type are provided in Online Resource~1. The primary defect categories are as follows.

\begin{itemize}
    \item \textbf{Background Artifacts and Low Contrast:} Aged, yellowed, or gray paper backgrounds diminish contrast between text and page (Fig.~\ref{fig:intro-page-2}; Online Resource~1).
    \item \textbf{Page Skew and Alignment Issues:} Many pages suffer from skew, where content is not aligned horizontally (Online Resource~1). This is a well-documented problem in \ac{ocr} literature~\cite{biswas2023document} that often requires specialized pre-processing to correct.
    \item \textbf{Text Bleed-Through:} On thin paper, ink from the reverse side is visible, creating superimposed text that interferes with primary content (Online Resource~1).
    \item \textbf{Water Damage:} Some documents show water damage resulting in blurred ink, stains, and overlapping text (Online Resource~1).
    \item \textbf{Physical Damage:} Prevalent physical damage includes tears, holes, and worn edges (Online Resource~1). This ranges from corner tears to significant edge damage and binding punch holes.
    \item \textbf{Stamps and Annotations:} Official stamps and ink annotations are frequently found, sometimes appearing as faint graphical elements (Online Resource~1). Only fillable stamps are of interest since they are processed as tabular data.
    \item \textbf{Scanning Artifacts from Bound Volumes:} Scanning from thick bound journals introduces page curl and a dark gradient near the inner margin (Online Resource~1).
    \item \textbf{Mixed Content:} Pages rarely contain a single content type; mixed layouts combining printed tables, handwritten notes, and drawings are common (Online Resource~1).
    \item \textbf{Manual Corrections:} Crossed-out words, removed paragraphs, and interlinear additions are common throughout the collection (Online Resource~1).
\end{itemize}

These complex and overlapping defects make naive document processing unreliable, underscoring the need for specialized classifiers robust to visual noise.

\begin{figure*}[htbp]\centering
\includegraphics[width=130mm]{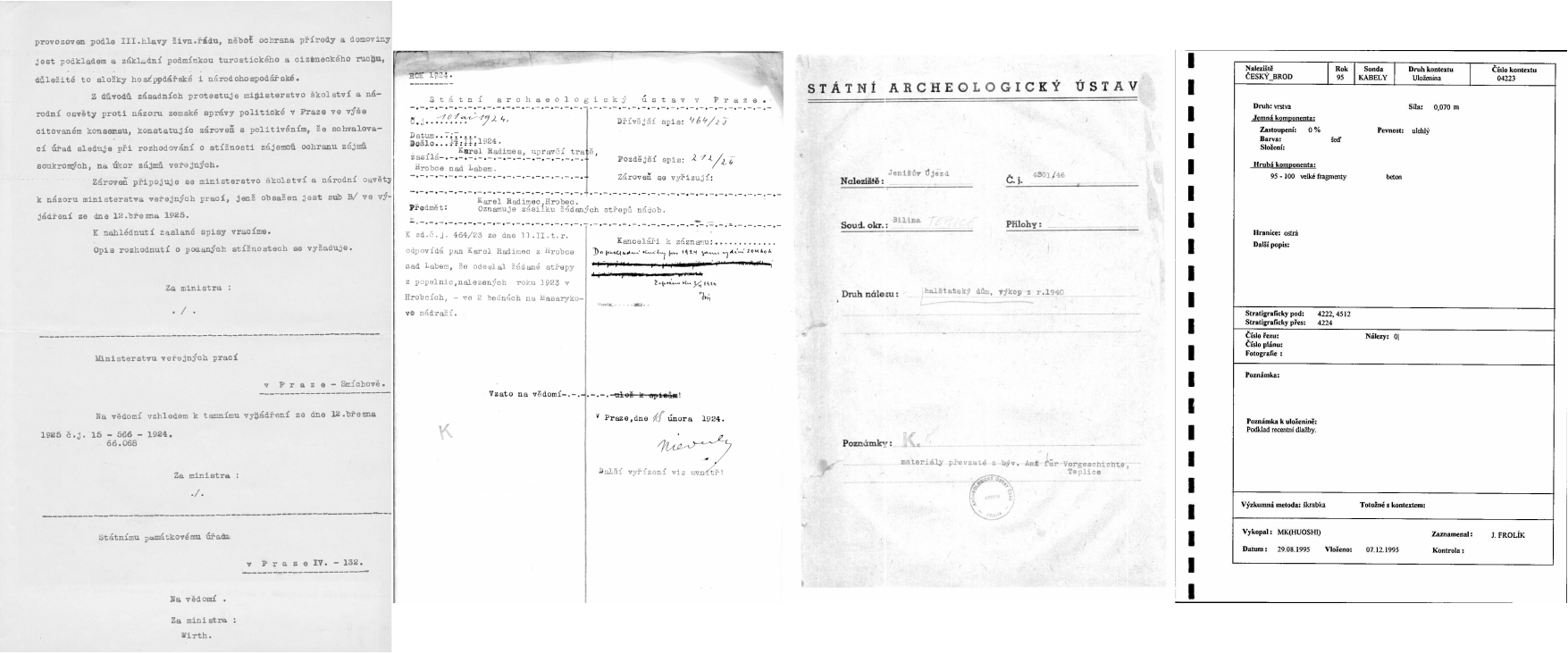}
\caption{Samples of pages with complex layouts that degrade standard OCR performance}
\label{fig:intro-ocr}
\end{figure*}

\subsubsection{Textual Variations and Annotations} \label{subsubsec212}

\noindent Beyond physical defects, the documents displayed considerable heterogeneity in text presentation and annotation. Pages frequently combined multiple text formats — for instance, typed documents with handwritten page numbers or comments (Online Resource~3). Textual elements were commonly embedded within graphical content; maps and drawings included labels and captions, while photographs were often accompanied by typed or handwritten descriptions (Online Resource~3). Consequently, purely graphical pages devoid of any text were relatively rare.

\section{Methods}\label{sec3}

\subsection{Preliminary Analysis and Justification} \label{subsec31}

\noindent To establish a performance baseline, an initial analysis was conducted using the \href{https://github.com/deepdoctection/deepdoctection}{DeepDoctection} \ac{dla} framework and the Tesseract \ac{ocr} engine~\cite{smith2007overview}. These experiments revealed that off-the-shelf tools were insufficient for the project's goals. \ac{ocr} performance degraded substantially on pages with noisy backgrounds (Online Resource~2), structured data detection was unreliable with frequent misclassification of tables (Online Resource~2), and graphical element detection was often inaccurate (Online Resource~2).

For \ac{dla} and table recognition, Detectron2 was used under the hood of \ac{dd}'s table recognition module. Table detection was unreliable: \ac{dd} often merged rows or missed cells when borders were faint, incomplete, or skewed (Online Resource~2). Maps were sometimes classified as tables, and handwritten annotations on drawings further confused the detector (Online Resource~2).

Feedback from a domain expert at \ac{arup} confirmed that out-of-the-box \ac{dla} performance was inadequate. Collaboration with the domain expert established a clear set of criteria for page classification, finalized through evaluation on a held-out set of documents:

\begin{itemize}
    \item \textbf{Classification Consistency:} Pages with similar content must receive the same category. Consistency was prioritized over isolated correctness.
    \item \textbf{Primacy of Structured Data:} Pages containing tabular or form-like data must be classified as such, even when significant plain text is also present.
    \item \textbf{Priority of Graphical Content:} A significant graphical element (at least stamp-sized) should take precedence over textual content in determining page category.
    \item \textbf{Handling of Handwritten Annotations:} Minor peripheral notes (e.g., page numbers) may be ignored; central or prominent handwritten elements should influence classification.
    \item \textbf{Robustness to Defects:} The system must be robust to background noise and paper degradation, which caused \ac{dd} to detect numerous spurious tables and figures (Online Resource~2).
\end{itemize}

This expert input finalized the category definitions and annotation guidelines. The first set of categories proposed by the data providers is described in Table~\ref{tab:categories_initial}.

\begin{table*}[htbp]
\centering
\begin{tabular}{l|p{10cm}}
\hline
\textbf{Category} & \textbf{Description} \\ \hline
REST & Mixture of printed, handwritten, and/or typed text, potentially with minor graphical elements (contained all ambiguous cases). \\ \hline
TEXT\_LINE & Pages primarily consisting of machine-typed, printed, or handwritten text organized in a tabular or form-like structure. \\ \hline
PHOTO & Pages dominated by photographs or photographic cutouts (including maps, paintings, schematics) with few text captions. \\ \hline
PHOTO\_TEXT & Similar to \texttt{PHOTO}, but the visual content is presented with a text block of any style. \\ \hline
TEXT & Pages containing plain corpora of almost pure printed, handwritten, or typed text. \\ \hline
TEXT\_OTHER & Pages containing mixtures of printed, handwritten, and/or typed text, potentially with minor graphical elements.
\end{tabular}
\caption{Overview of categories initially proposed by the data provider}
\label{tab:categories_initial}
\end{table*}

\subsection{Supervised Image Classification Models} \label{subsec32}

\noindent Given the limitations of unsupervised \ac{dla} methods on heterogeneous historical document data, our focus shifted to supervised image classification. This section details the evolution of classification categories, the development of a baseline low-compute model, the evaluation of state-of-the-art deep learning architectures, and the iterative data refinement that was crucial for achieving high accuracy.

Initially, the categories were defined based on the capabilities of the \ac{dla} framework we first tested (Table~\ref{tab:categories_initial}). We then adjusted this scheme to better align with the capabilities of statistical models, noting that a distinction between handwritten and other text types was a promising direction. This led to a proposed set of 7 refined categories (Table~\ref{tab:categories_proposal}).

\begin{table*}[htbp]
\centering
\begin{tabular}{r|p{10cm}}
\hline
\textbf{Category} & \textbf{Description} \\ \hline
DRAW\_TEXT & Pages dominated by drawings, maps, paintings, schematics, or graphics with text labels. \\ \hline
TEXT\_LINE & Pages primarily consisting of machine-typed, printed, or handwritten text organized in a tabular or form-like structure. \\ \hline
PHOTO & Pages dominated by drawings, maps, paintings, schematics, graphics, photographs, or photographic cutouts, with few text captions. \\ \hline
PHOTO\_TEXT & Similar to \texttt{PHOTO}, but with a significant text block. \\ \hline
TEXT & Pages containing plain corpora of almost pure printed or typewritten text. \\ \hline
HW & Pages consisting purely of handwritten text in paragraph or block form (non-tabular). \\ \hline
TEXT\_HW & Pages containing mixtures of handwritten and typed text, potentially with minor graphical elements.
\end{tabular}
\caption{Overview of categories as modifications of the initial ones}
\label{tab:categories_proposal}
\end{table*}

After demonstrating initial results with these categories, the data providers agreed to expand the label set. Following expert feedback, the \texttt{PHOTO\_TEXT} category was replaced by \texttt{PHOTO\_L} and \texttt{DRAW\_L} to better distinguish graphical content in tabular layouts, and the \texttt{TEXT} and \texttt{TEXT\_LINES} categories were subdivided into handwritten (\texttt{\_HW}), machine-typed (\texttt{\_T}), and printed (\texttt{\_P}) variants.

\begin{figure*}[htbp]
    \centering
    \caption{Distribution of categories based on document creation year (final annotated training dataset)}
    \label{fig:label_time}
        \includegraphics[width=\textwidth]{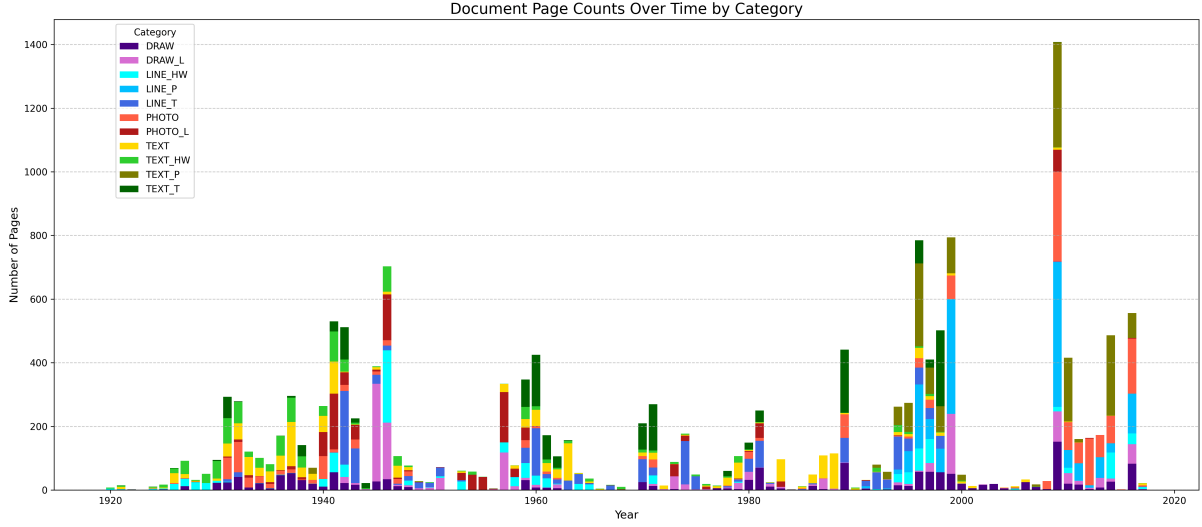}
\end{figure*}

This collaborative process ultimately produced a final set of 11 distinct categories (Table~\ref{tab:categories_used}). Category-specific temporal characteristics of the final annotation are plotted in Fig.~\ref{fig:label_time}. Figure~\ref{fig:label_time} also reveals a gap around the 1990s, when printed monospaced fonts became visually indistinguishable from typewritten text. Pages from this period were not removed from the source collection; rather, they were left unannotated and thus excluded from the ground truth, letting the models infer the distinction on their own. Because these pages were never assigned a label, they are not enumerated in the annotation counts of Table~\ref{tab:category-distribution-datasets} and appear only as the visible trough around 1990 in Fig.~\ref{fig:label_time}.

\subsubsection{Classification Categories and Priorities}
\label{subsubsec311}

\noindent The 11 target classes (Table~\ref{tab:categories_used}) cover almost half of all content type combinations (Table~\ref{tab:11-label} versus Table~\ref{tab:24-label} in Appendix~\ref{secC1}). Visual examples for each category are provided in Online Resource~3. Annotation was carried out jointly by the first author and a domain-expert end-user from \ac{arup}, using the priority order below as the shared annotation guideline.

\begin{table*}[htbp]
\centering
\begin{tabular}{r|p{10cm}}
\hline
\textbf{Category} & \textbf{Description} \\ \hline
DRAW & Pages dominated by drawings, maps, paintings, schematics, or graphics, potentially containing text labels or captions. \\ \hline
DRAW\_L & Similar to \texttt{DRAW}, but the graphical element is presented within a table-like layout or includes a legend formatted as a table. \\ \hline
LINE\_HW & Pages primarily consisting of handwritten text organized in a tabular or form-like structure. \\ \hline
LINE\_P & Pages primarily consisting of printed text organized in a tabular or form-like structure. \\ \hline
LINE\_T & Pages primarily consisting of machine-typed text organized in a tabular or form-like structure. \\ \hline
PHOTO & Pages dominated by photographs or photographic cutouts, potentially with text captions. \\ \hline
PHOTO\_L & Similar to \texttt{PHOTO}, but the photograph is presented within a table-like layout or accompanied by tabular annotations. \\ \hline
TEXT & Pages containing mixtures of printed, handwritten, and/or typed text, potentially with minor graphical elements. \\ \hline
TEXT\_HW & Pages consisting purely of handwritten text in paragraph or block form (non-tabular). \\ \hline
TEXT\_P & Pages consisting purely of printed text in paragraph or block form (non-tabular). \\ \hline
TEXT\_T & Pages consisting purely of machine-typed text in paragraph or block form (non-tabular).
\end{tabular}
\caption{Overview of categories used in the trained models}
\label{tab:categories_used}
\end{table*}

A priority order was established to handle pages potentially fitting multiple categories:
\begin{enumerate}
    \item PHOTOs (\texttt{PHOTO}, \texttt{PHOTO\_L}): Highest priority for graphic extraction.
    \item DRAWs (\texttt{DRAW}, \texttt{DRAW\_L}): Second priority for graphic extraction.
    \item LINEs (\texttt{LINE\_HW}, \texttt{LINE\_P}, \texttt{LINE\_T}): Third priority for structured data extraction.
    \item TEXTs (\texttt{TEXT}, \texttt{TEXT\_HW}, \texttt{TEXT\_P}, \texttt{TEXT\_T}): Lowest priority, targeting font-specific \ac{ocr}.
\end{enumerate}

To capture the full variability of the data, we also created an expanded label scheme of 24 distinct types by separating each core category into printed, typewritten, and handwritten variants (Table~\ref{tab:24-label}). This comprehensive set was used for analytical purposes, while the 11-category set was used for training the final models.

\paragraph{Annotation reliability.} The ground truth was produced by two annotators --- the first author and a domain-expert end-user from \ac{arup} (Ing.\ Dana Křivánková, the fourth author) --- under the oversight of the project lead (Mgr.\ David Novák, Ph.D.). Annotation followed an iterative, expert-reviewed protocol: the first author labelled batches that the domain expert then reviewed and approved, and the label disagreements surfaced in this loop directly drove the evolution of the scheme from 6 to 7 to the final 11 categories (Tables~\ref{tab:categories_initial}--\ref{tab:categories_used}). The fixed priority rule above (graphics $>$ structured data $>$ text) was the shared mechanism for resolving multi-content pages to a single label. Because every page ultimately carries one adjudicated consensus label, a conventional inter-annotator agreement coefficient is not defined over the full released dataset, and we did not compute a formal Cohen's~$\kappa$; the authors' qualitative impression is that batch-level agreement improved from roughly $\kappa\approx0.5$ early in the project to $\kappa\approx0.9$ once the taxonomy and guidelines stabilized. A controlled dual-annotation study on a held-out subset to report a formal $\kappa$ is left for future work. In the interim, reliability is evidenced indirectly by the explicit priority rules, the iterative expert review across the four dataset versions (Sect.~\ref{subsubsec324}), and the high cross-fold stability and inter-model agreement of the resulting classifiers (Sects.~\ref{subsec52} and~\ref{subsec43}).

\subsubsection{Low-Compute Baseline: Random Forest Classifier} \label{subsubsec321}

\begin{figure*}[htbp]
    \centering
    \caption{Comparison of low-compute classifiers (RFC, LDA, k-NN, SVM, Naive Bayes, logistic regression) on the data-provider scheme (left) and our proposed scheme (right), evaluated via cross-fold validation on fewer than 2,000 images. RFC outperforms all classical alternatives on both label sets}
    \label{fig:low-compared}
        \includegraphics[width=\textwidth]{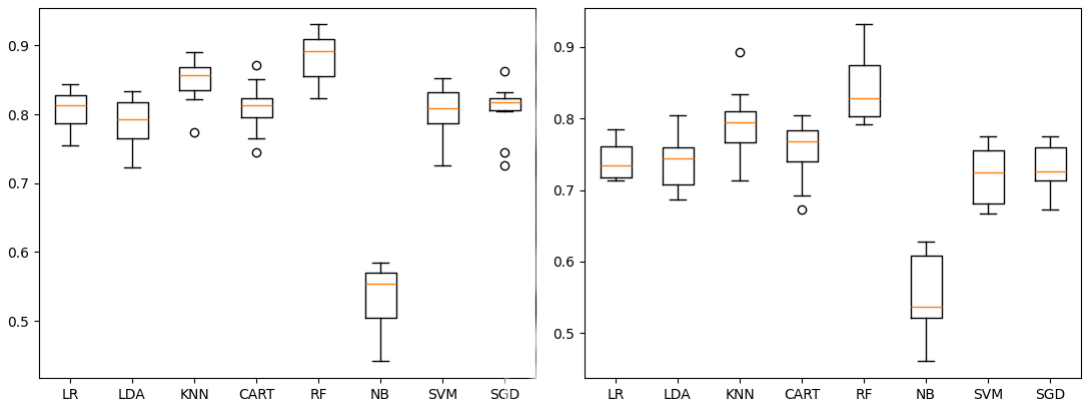}
\end{figure*}

\noindent As a computationally efficient baseline, we implemented a traditional computer vision pipeline using a \acf{rfc}~\cite{breiman2001random} requiring no specialized \acp{gpu}. Each image was transformed into a numerical feature vector combining descriptors from grayscale and binarized versions:
\begin{itemize}
    \item \textbf{Preprocessing:} Grayscale conversion, Otsu thresholding~\cite{yousefi2011image}, and basic image properties (dimensions, pixel ratios).
    \item \textbf{Hu Moments~\cite{hu1962visual}:} Seven invariant moments capturing shape information robust to translation, scale, and rotation.
    \item \textbf{Haralick Texture Features~\cite{haralick1973textural}:} Statistics derived from Gray-Level Co-occurrence Matrices (GLCM), describing contrast, correlation, and homogeneity.
    \item \textbf{Histogram Features:} Pixel intensity distributions (256 bins for grayscale, 2 bins for binary).
\end{itemize}
The resulting 298-float representation enables efficient processing on modest hardware. Several classical classifiers were evaluated on this feature vector — including \ac{lda}, \ac{knn}, Naive Bayes, \ac{svm}, and logistic regression — with \ac{rfc} achieving the best preliminary accuracy, as shown in Fig.~\ref{fig:low-compared}. The \ac{rfc} was therefore selected for its efficiency and interpretability. However, this approach achieved only $\sim$75\% accuracy, demonstrating the need for more powerful deep learning models.

\begin{figure*}[htbp]
    \centering
    \caption{RFC experiments on data providers' (left) and ours (right) annotations of fewer than 2,000 images}
    \label{fig:rfc}
        \includegraphics[width=\textwidth]{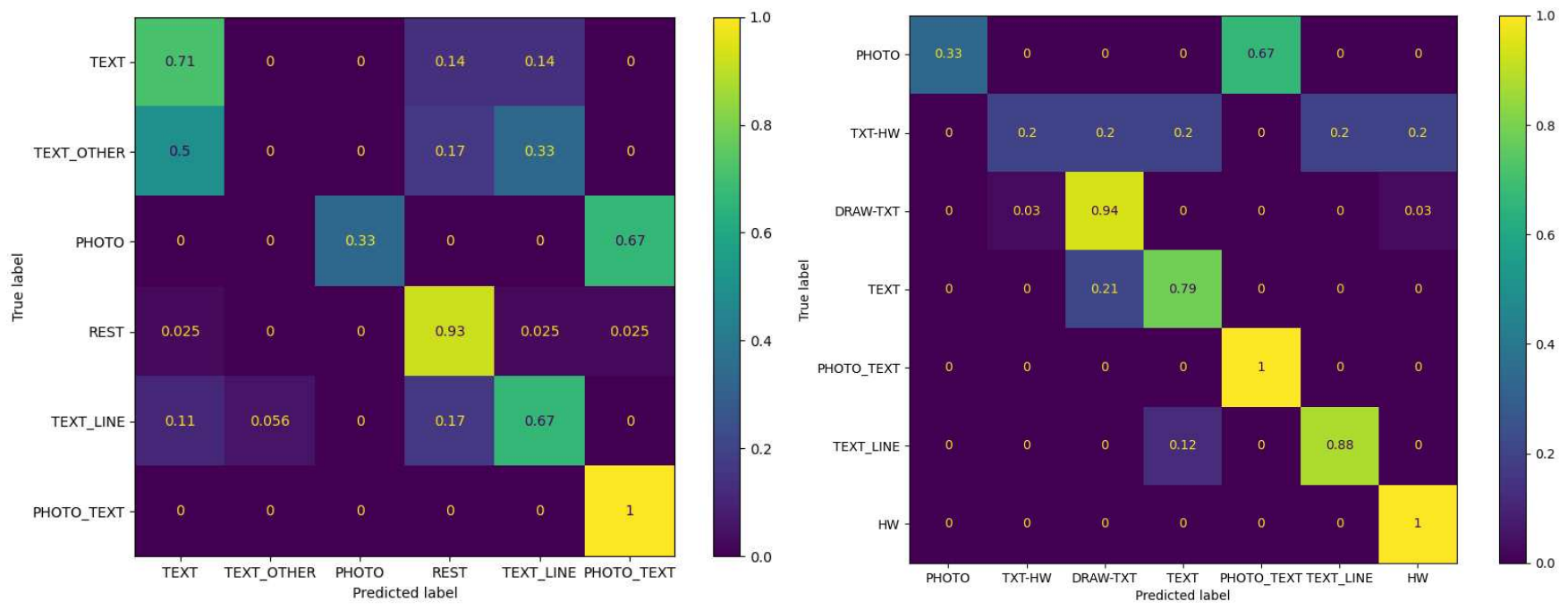}
\end{figure*}

At that point, we annotated data according to the data provider suggestions (Table~\ref{tab:categories_initial}) and our own refined propositions (Table~\ref{tab:categories_proposal}), then trained the \ac{rfc} model to demonstrate its performance at the first workshop. Development-set evaluation results are shown in Fig.~\ref{fig:rfc}.

\subsubsection{Deep Learning Approaches: CNNs and Transformers} \label{subsubsec322}

\noindent We evaluated several deep neural network architectures by fine-tuning models pre-trained on different datasets. All models in Table~\ref{tab:models} were fine-tuned with a consistent set of data augmentations, using five-fold cross-validation with an 80\%/10\%/10\% train/development/test split on the final annotated dataset. Image-only models were fine-tuned for 3~epochs per fold; the resulting per-fold checkpoints were subsequently averaged to produce the final deployment weights, following the procedure described in~\cite{lutsai2026pageimageclassification}.

\textbf{Training configuration.} All image-only models were trained using AdamW with a learning rate of $10^{-5}$, linear warmup over 10\% of training steps, and per-epoch checkpointing with best-model selection based on development-set accuracy. Batch size was set to 12--16 pages depending on available GPU memory. Data augmentation applied randomly and independently per image: brightness, contrast, saturation, and hue adjustments (each at probability~0.5), sharpness scaling (random factor 0.5--1.5), and Gaussian blur (random radius 0--2). Geometric transformations (rotation, flipping) were deliberately excluded to preserve document orientation cues.

\begin{table*}[htbp]
\centering
\begin{tabular}{@{}lcccc@{}}
\toprule
Model & Input Size & Pre-training Dataset & Params (M) \\ \midrule
EfficientNet-v2-S & 300$\times$300 & ImageNet-21k & 48.2 \\
EfficientNet-v2-M & 384$\times$384 & ImageNet-21k \& ImageNet-1k & 54.1 \\
EfficientNet-v2-L & 384$\times$384 & ImageNet-21k \& ImageNet-1k & 118.5 \\ \midrule
RegNetY-12GF & 224$\times$224 & ImageNet-12k \& ImageNet-1k & 51.8 \\
RegNetY-16GF & 224$\times$224 & ImageNet-12k & 83.6 \\
RegNetY-64GF & 384$\times$384 & SEER \& ImageNet-1k & 281.4 \\ \midrule
dit-base-rvlcdip & 224$\times$224 & IIT-CDIP (42M) \& RVL-CDIP & 86 \\
dit-large & 224$\times$224 & IIT-CDIP (42M) & 304 \\
dit-large-rvlcdip & 224$\times$224 & IIT-CDIP (42M) \& RVL-CDIP & 304 \\ \midrule
vit-base-patch16 & 224$\times$224 & ImageNet-21k \& ImageNet-1k & 86.6 \\
vit-base-patch16 & 384$\times$384 & ImageNet-21k \& ImageNet-1k & 86.9 \\
vit-large-patch16 & 384$\times$384 & ImageNet-21k \& ImageNet-1k & 304.7 \\ \midrule
CLIP-ViT-B/32 & 224$\times$224 & WebImageText (400M) & 151 \\
CLIP-ViT-B/16 & 224$\times$224 & WebImageText (400M) & 150 \\
CLIP-ViT-L/14 & 224$\times$224 & WebImageText (400M) & 428 \\
CLIP-ViT-L/14@336px & 336$\times$336 & WebImageText (400M) & 428 \\ \bottomrule
\end{tabular}
\caption{Specifications of all evaluated models}
\label{tab:models}
\end{table*}

\begin{itemize}
    \item \textbf{\acp{cnn} (EfficientNetV2 and RegNetY):} We fine-tuned variants of EfficientNetV2~\cite{tan2021efficientnetv2,tan2019efficientnet} and RegNetY~\cite{radosavovic2020designing}. The final classification layer was replaced to match our 11 classes. Among EfficientNetV2, the medium variant proved most efficient at 98.83\%; RegNetY-16GF emerged as the top overall performer at 99.16\%, surpassing its larger 64GF sibling.
    \item \textbf{Vision Transformers (\ac{vit} and \ac{dit}):} We fine-tuned variants of \acf{vit}~\cite{dosovitskiy2020image,beyer2022better} and \acf{dit}~\cite{li2022dit}, the latter pre-trained on large-scale document images from IIT-CDIP~\cite{lewis2006building} and optionally fine-tuned on RVL-CDIP~\cite{harley2015icdar}. ViT-large reached $\approx$99.12\% Top-1 accuracy; DiT variants achieved 98--99\%, consistent with \acp{cnn} but not surpassing the best CNN models.
\end{itemize}

\subsubsection{Multimodal Approach: CLIP} \label{subsubsec323}

\noindent Finally, we investigated a multimodal approach using \acf{clip}~\cite{radford2021learning}, which learns visual representations from natural language supervision. In contrast to image-only models, \ac{clip} was fine-tuned for 7~epochs with a learning rate of $5\times10^{-5}$ on the first fold of the cross-validation split. This asymmetric evaluation (one fold rather than five) was a pragmatic choice given the substantially higher computational cost of \ac{clip} fine-tuning; results should therefore be interpreted as indicative rather than fully comparable to the five-fold image-only scores.

\begin{figure*}[htbp]
\centering
\includegraphics[width=130mm]{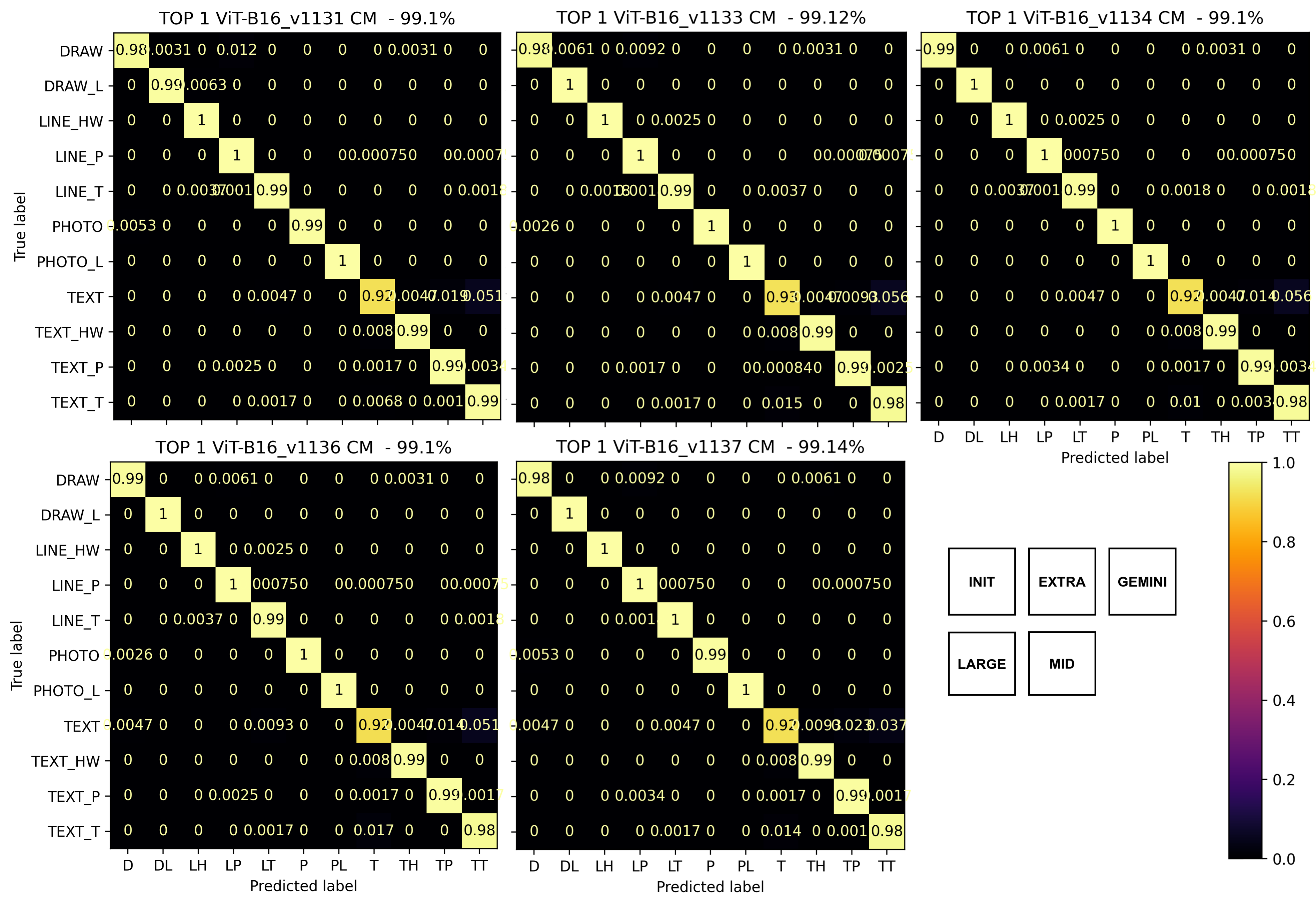}
\caption{Confusion matrix of the best CLIP configuration: ViT-B/16 fine-tuned with the \textit{mid} category description set (Table~\ref{tab:short_classification}), achieving 99.14\% accuracy on the standard test set (5,449~pages)}
\label{fig:clip-best5}
\end{figure*}

The best-performing configuration — CLIP-ViT-B/16 with the \textit{mid} category description set (Table~\ref{tab:short_classification}) — reached 99.14\% accuracy on the held-out test set (Fig.~\ref{fig:clip-best5}), comparable to the best image-only models. The larger ViT-L variants scored approximately 98.97\% (see~\cite{lutsai2026pageimageclassification} for full confusion matrices of all CLIP variants).

\begin{figure*}[htbp]
\centering
\includegraphics[width=\textwidth]{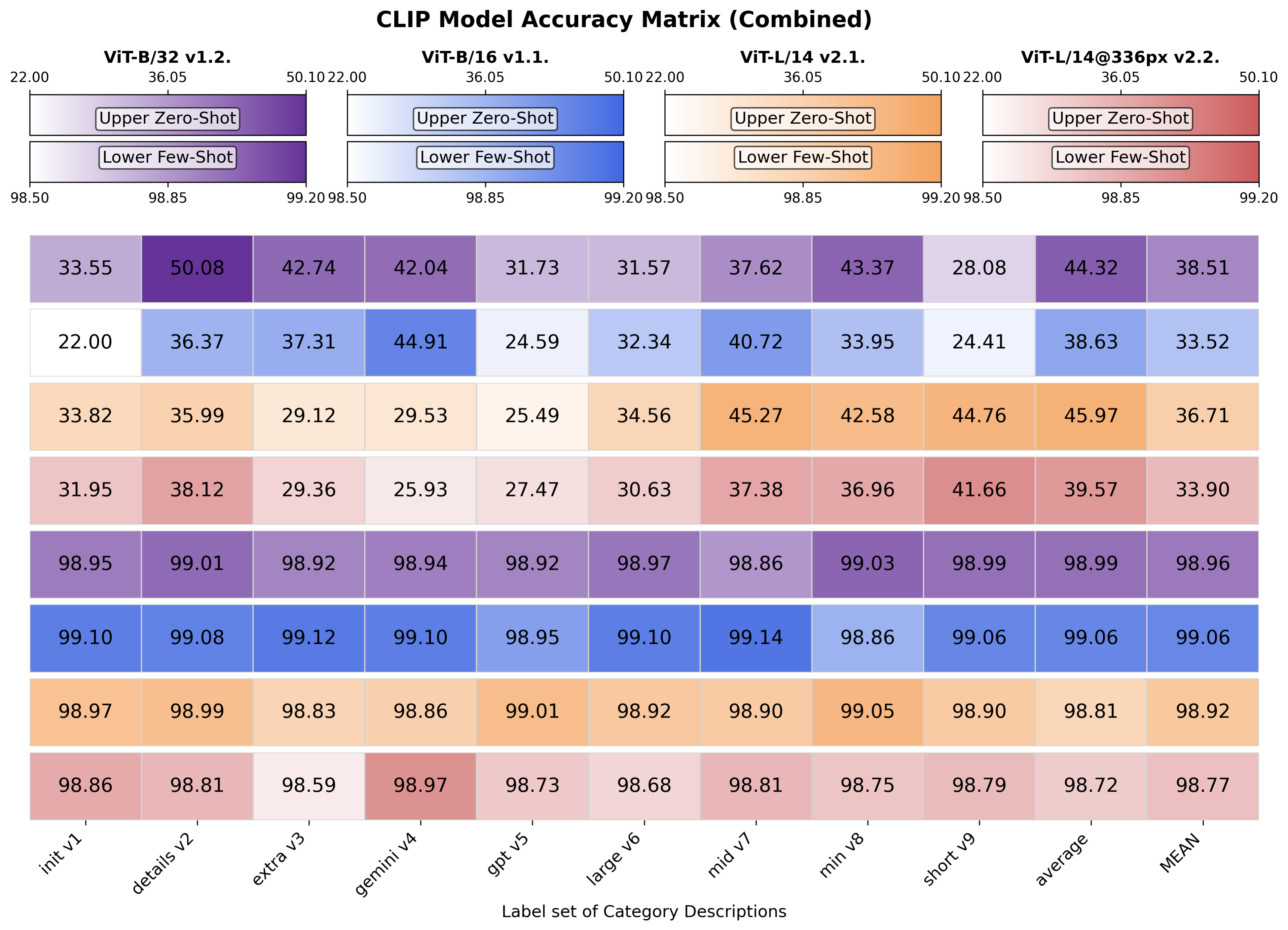}
\caption{Combined zero-shot and fine-tuned \ac{clip} models: comparison of classification accuracy across category description sets on the standard test dataset (5{,}449 pages)}
\label{fig1:clip-acc-matrix}
\end{figure*}

\begin{figure*}[htbp]
\centering
\includegraphics[width=\textwidth]{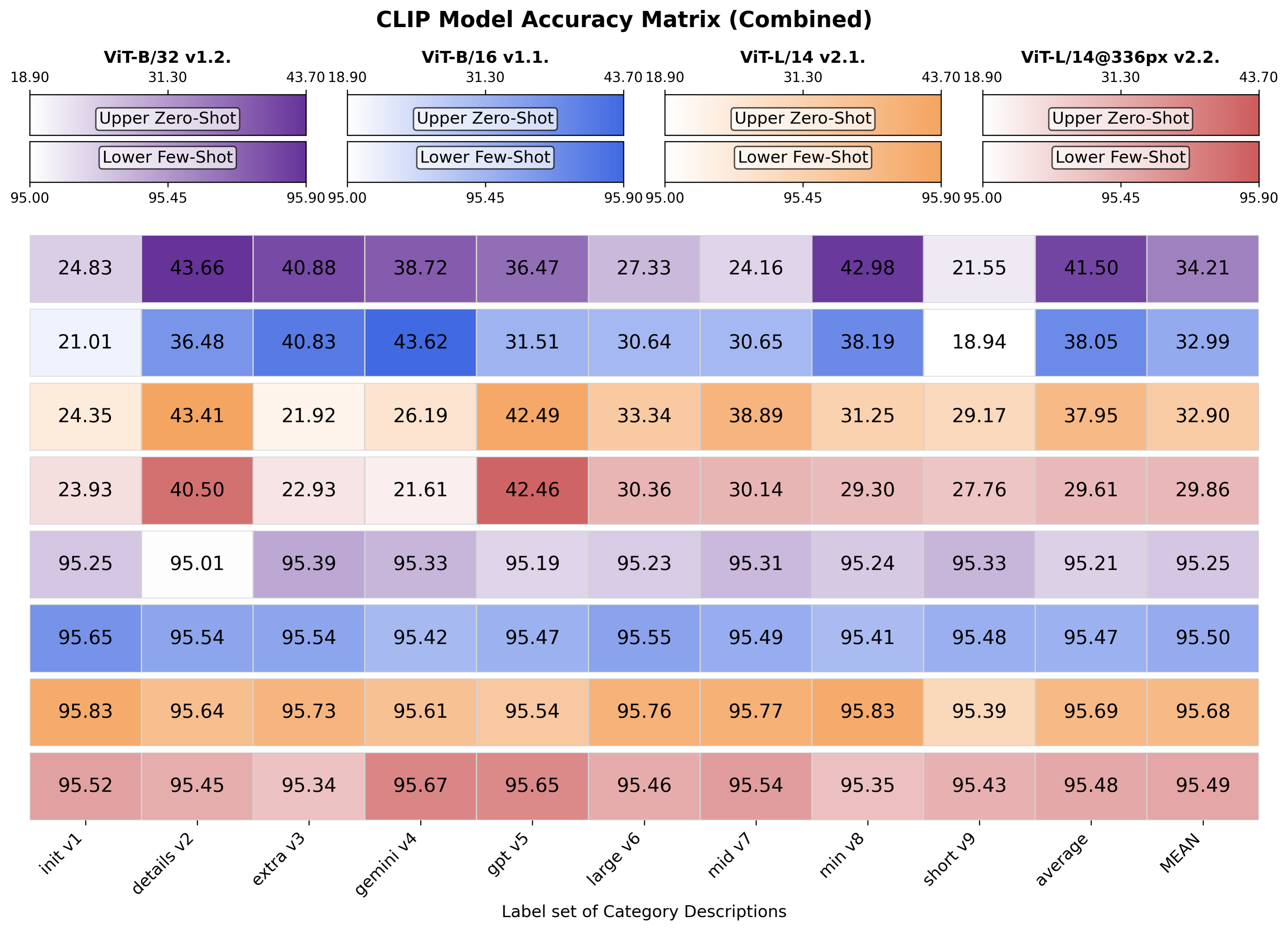}
\caption{Combined zero-shot and fine-tuned \ac{clip} models: comparison of classification accuracy across category description sets on an expanded test dataset (all non-training samples from the first-fold split, random seed 420; 14{,}162 pages)}
\label{fig1:clip-acc-matrix-big}
\end{figure*}

\textbf{Category descriptions.} The classification task was tested using predefined sets of categories, each paired with a distinct textual description to guide the \ac{clip} model. A summary of all description sets is given in Table~\ref{tab:summary_classification} in Appendix~\ref{secA1}; representative examples are provided in Tables~\ref{tab:init_classification} and~\ref{tab:short_classification}. The full suite of eight description sets, ranging from minimalist to detailed approaches, is reproduced in full in the accompanying thesis~\cite{lutsai2026pageimageclassification}.

Initial experiments revealed varying error rates across individual description sets, leading to an averaging approach that leverages all descriptions to capture a more generalized category representation. Both zero-shot and fine-tuned results were derived from the precomputed mean of all category descriptions' text features. Forward-looking implications of \ac{clip}'s real-world deployment behavior are discussed in Sect.~\ref{subsec43}.

\subsubsection{Dataset Curation and Refinement} \label{subsubsec324}

\noindent Achieving high accuracy was not just a matter of model selection but also of careful dataset curation. The training data evolved through four successive annotation stages, as summarized in Table~\ref{tab:category-distribution-datasets}. We measured the impact of these changes by evaluating a baseline \ac{vit} model on each dataset version.

\begin{table*}[htbp]
\centering
\caption{Category distribution across annotation versions. Datasets~0--2 share the same 11-category scheme but use earlier annotation passes; the full per-category counts for all four versions are listed. The \texttt{TEXT} category was capped at 14,000 samples for training (227 additional \texttt{TEXT} pages were reserved for test evaluation only). Dataset~1 (\textit{poor-selection}) was abandoned after evaluation showed removing noisy-but-distinct pages was detrimental}
\label{tab:category-distribution-datasets}
\begin{tabular}{r|rrrr}
\hline
\textbf{Category} & \textbf{Dataset 0} & \textbf{Dataset 1} & \textbf{Dataset 2} & \textbf{Dataset 3} \\ \hline
DRAW      & 1,090 & 1,368 & 1,472 & 2,709 \\
DRAW\_L   & 1,091 & 1,383 & 1,402 & 2,921 \\
LINE\_HW  & 1,055 & 1,113 & 1,115 & 2,514 \\
LINE\_P   & 1,092 & 1,540 & 1,580 & 2,439 \\
LINE\_T   & 1,098 & 1,664 & 1,668 & 9,883 \\
PHOTO     & 1,081 & 1,632 & 1,730 & 2,691 \\
PHOTO\_L  & 1,087 & 1,087 & 1,088 & 2,830 \\
TEXT      & 1,091 & 1,587 & 1,592 & 14,227 \\
TEXT\_HW  & 1,091 & 1,092 & 1,092 & 2,008 \\
TEXT\_P   & 1,083 & 1,540 & 1,633 & 2,312 \\
TEXT\_T   & 1,081 & 1,476 & 1,482 & 3,965 \\ \hline
\textbf{Unique PDFs} & 5,001 & 5,694 & 5,729 & 37,328 \\
\textbf{Total Pages} & \textbf{11,940} & \textbf{15,482} & \textbf{15,854} & \textbf{48,499} \\ \hline
\end{tabular}
\end{table*}

Initially, our annotation set (Dataset~0: 11,940 pages) had limited document variety. We expanded samples within each category, corrected misclassifications, and reclassified ambiguous items like stamps. One iteration, referred to as \textit{poor selection} (Dataset~1), involved removing samples that seemed too noisy or redundant; however, evaluation revealed that removing these distinct examples was detrimental to performance. Consequently, the final annotation phase (Dataset~3: 48,499 pages) involved restoring removed pages to ensure sufficient sample diversity, particularly for the \texttt{PHOTO}, \texttt{TEXT\_P}, and \texttt{TEXT\_T} categories.

The pronounced class imbalance in Dataset~3 is not a design choice but reflects the natural composition of the archive, in which mixed-style plain text (\texttt{TEXT}) and typed tabular pages (\texttt{LINE\_T}) dominate. To prevent the majority \texttt{TEXT} class (14,227 available pages) from overwhelming training and biasing the model against the minority graphical and tabular categories, we capped \texttt{TEXT} at 14,000 pages for the training/development/test partition; the 227 surplus \texttt{TEXT} pages were not discarded but reserved for test-only evaluation, so that measured accuracy on \texttt{TEXT} remains unbiased by the cap. We retained the natural imbalance of the remaining categories and addressed it at training time with a \texttt{BalancedBatchSampler} (Sect.~\ref{subsec33}) rather than by further sub-sampling, since the \textit{poor-selection} experiment had shown that discarding distinct pages hurts generalization.

\begin{figure*}[htbp]
\centering
\includegraphics[width=130mm]{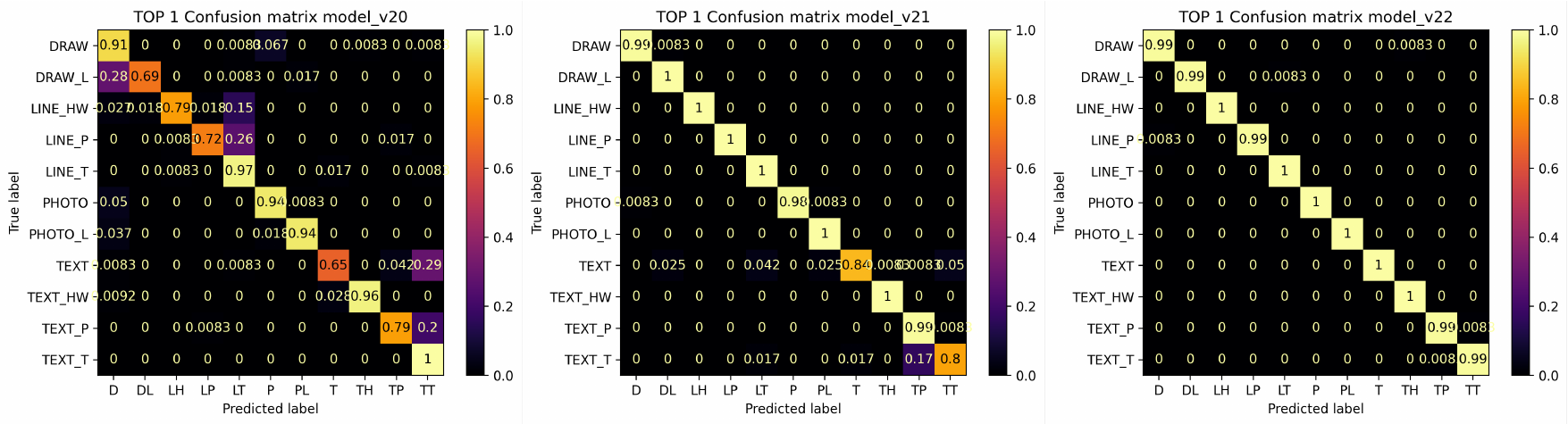}
\caption{Confusion matrices of Top-1 predictions for ViT-Base 224$\times$224 fine-tuned on successive dataset versions. From left to right: Dataset~0 (initial, 11,940 pages), Dataset~1 (poor selection, 15,482 pages), Dataset~3 (final, 48,499 pages)}
\label{fig:vits}
\end{figure*}

Figure~\ref{fig:vits} shows the Top-1 prediction confusion matrices for ViT-Base trained on the initial, poor-selection, and final dataset versions, illustrating performance improvements gained through careful data curation. Temporal category distributions across all three annotation versions, revealing the characteristic 1990s gap where printed monospaced fonts were excluded, are provided in Online Resource~4.

For the final Dataset~3 models, we used a five-fold cross-validation scheme with an \textbf{80\%}/\textbf{10\%}/\textbf{10\%} train/development/test split. For each fold the final subset counts were 38,625~/~4,823~/~4,823 for train/development/test respectively. Rather than a naive random shuffle, a time-aware periodic sampling procedure was employed: for each category, every $S$-th element is drawn from the alphabetically (approximately chronologically) ordered sequence, with a bounded random offset, ensuring chronological coverage and reducing the risk of evaluation sets dominated by a single scanning campaign~\cite{lutsai2026pageimageclassification}. Consistent category coverage across all five folds is confirmed by Online Resource~4.

Overall, 43,050 images were used across the five training subsets; the remaining 5,449 images formed the final performance test subset used for all model comparisons (Sect.~\ref{subsec52}).

\subsection{System Architecture and Implementation} \label{subsec33}

\noindent The system is implemented as a modular Python application using PyTorch~\cite{paszke2019pytorch} and Hugging Face's Transformers library~\cite{wolf2019huggingface}. The complete codebase is publicly available in a GitHub repository~\cite{Lutsai_ATRIUM_s_page_classifier_2025} under the MIT license. The visualization of system architecture for various models is given in Figs.~\ref{fig:architect_trans} and~\ref{fig:architect_clip} in Appendix~\ref{secC1}.

\begin{table*}[htbp]
\centering
\caption{Configuration settings in \texttt{config.txt}}
\label{tab:config-settings}
\begin{tabular}{lp{11cm}}
\toprule
\textbf{Section} & \textbf{Description} \\
\midrule
\texttt{[INPUT]} & Specifies the default directory of input images for processing. \\
\texttt{[OUTPUT]} & Defines paths for result \ac{csv} files, model checkpoints, and visualization outputs. \\
\texttt{[SETUP]} & Contains operational parameters such as batch size and the value N for top-N predictions. \\
\texttt{[TRAIN]} & Stores training-specific settings: dataset path, number of epochs, learning rate, validation split, and logging frequency. \\
\texttt{[MODEL]} & Selects the model architecture (e.g., \ac{vit}, EfficientNet), version, and pre-trained weights. \\
\texttt{[EVAL]} & Specifies the default directory path for the evaluation dataset. \\
\texttt{[HF]} & Stores settings for Hugging Face integration and model repository management. \\
\bottomrule
\end{tabular}
\end{table*}

The system offers a \ac{cli} (\texttt{run.py}) for direct control over operations like training, evaluation, and inference (documented in Table~\ref{tab:cli-options} in Appendix~\ref{secC1}), along with a central configuration file (\texttt{config.txt}) for managing parameters such as model selection, learning rates, and input/output paths without code modification.

The system accepts single images or directories of images in standard formats (\ac{png}, \ac{jpeg}, etc.). For training, it employs a custom \textbf{BalancedBatchSampler} to ensure each batch contains a balanced representation of classes, improving training stability with imbalanced datasets. Data augmentation (brightness, contrast, saturation adjustments) is applied during training to improve model robustness; geometric transformations are avoided to preserve document layouts. The dataset is split into training (80\%), development (10\%), and test (10\%) sets using stratified sampling.

The system generates results in user-friendly formats including console output for single files, \ac{csv} tables with Top-N predictions for batch processing, and confusion matrix plots for performance visualization. Cross-platform (Linux/Windows) scripts are provided for converting multi-page \ac{pdf} documents into individual page images and for sorting annotated images into the required directory structure.

\begin{figure*}[htbp]
\centering
\includegraphics[width=130mm]{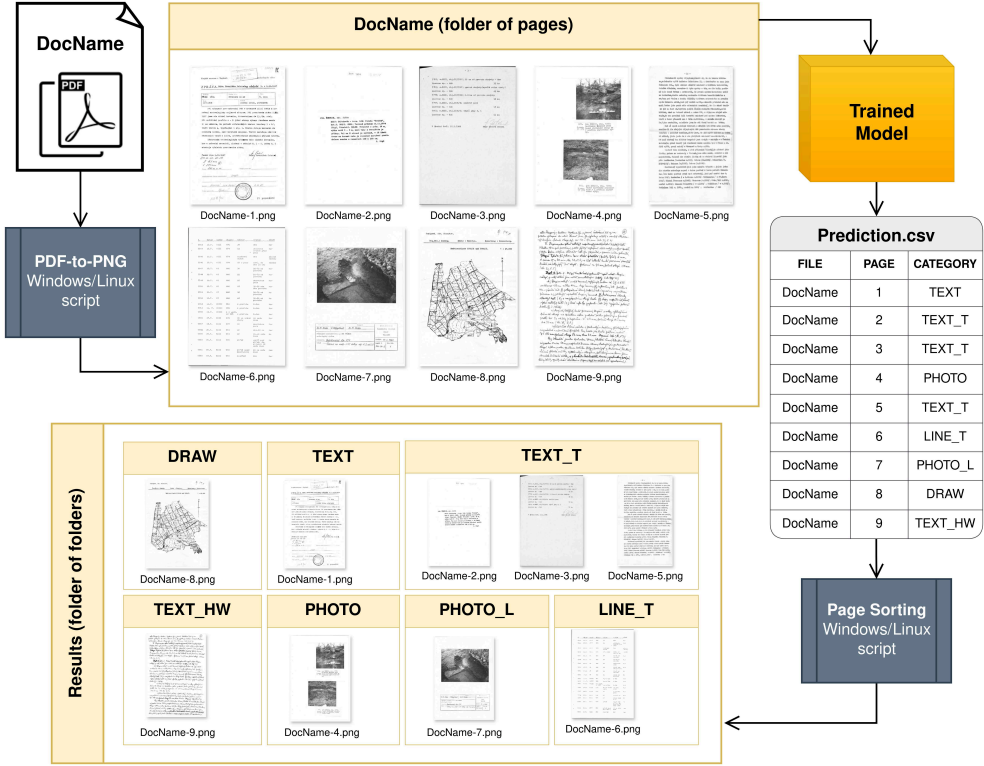}
\caption{Model inference use-case: processing a single PDF file into sorted and classified pages}
\label{fig:sorting}
\end{figure*}

\section{Results}\label{sec4}

\noindent This section details the practical integration of the classification systems into archival workflows and presents a comprehensive performance evaluation.

\subsection{Integration into Archives} \label{subsec51}

\noindent A primary objective was to develop a system that is not only accurate but also practical for use within existing archival processes.

\subsubsection{Deployment and Usability} \label{subsubsec511}

\noindent The system has been packaged for local deployment under the MIT license, ensuring suitability for any institutional environment. It is designed for on-premises operation, an essential feature for archives managing sensitive or restricted collections. Standard \ac{cpu} hardware suffices for inference, though a \ac{gpu} is recommended for fine-tuning on user-annotated datasets and for accelerating large-batch inference.

The core functionality is platform-agnostic, compatible with both Linux and Windows operating systems. On a modern office desktop computer, the system can process several hundred thousand pages in under a week, with throughput potentially exceeding one million pages on higher-end hardware. For optimal performance utilizing \ac{cuda} for \ac{gpu} acceleration, an NVIDIA graphics card is required. Users must allocate up to 30~GB disk space to accommodate Python dependencies, model weights, and datasets.

In a practical scenario, the classifier is integrated into the digitization pipeline: as documents are scanned, the tool assigns a classification label to each page, enabling efficient routing to subsequent processing stages — plain text to an \ac{ocr} engine, tables to a structured data extraction module, and illustrations to specialized image analysis.

\subsubsection{Agreement with Field Experts} \label{subsubsec512}

\noindent The system's classification logic was refined based on criteria established in collaboration with archival experts. The final expert-validated criteria are as follows:

\begin{itemize}
    \item All photographs or drawings of a significant size (at least postage-stamp sized) must be identified as \texttt{PHOTO} or \texttt{DRAW}, enabling targeted extraction of graphical content.
    \item All tables and forms, regardless of border visibility, must be detected for subsequent structured data extraction, classified as \texttt{LINE\_T}, \texttt{LINE\_P}, or \texttt{LINE\_HW}.
    \item Only pages containing clean, uniform text should be classified as \texttt{TEXT\_T}, \texttt{TEXT\_P}, or \texttt{TEXT\_HW}. Minor elements such as handwritten page numbers may be ignored.
    \item Pages with mixed text styles, classified as \texttt{TEXT}, may include minor graphical elements such as newspaper logos. Extraction of these small elements is deemed unnecessary.
    \item Graphical elements within tabular layouts or accompanied by table-like legends must be classified as \texttt{PHOTO\_L} or \texttt{DRAW\_L}, signaling the need to extract both graphical content and structured data.
\end{itemize}

\subsection{Accuracy Performance of Tested Models} \label{subsec52}

\begin{figure*}[htbp]
    \centering
    \caption{Distribution of categories in the performance test subset (5,449 pages not included in any cross-validation training fold), based on document creation year. The test set covers the full chronological range of the archive}
    \label{fig:label_time_test}
        \includegraphics[width=\textwidth]{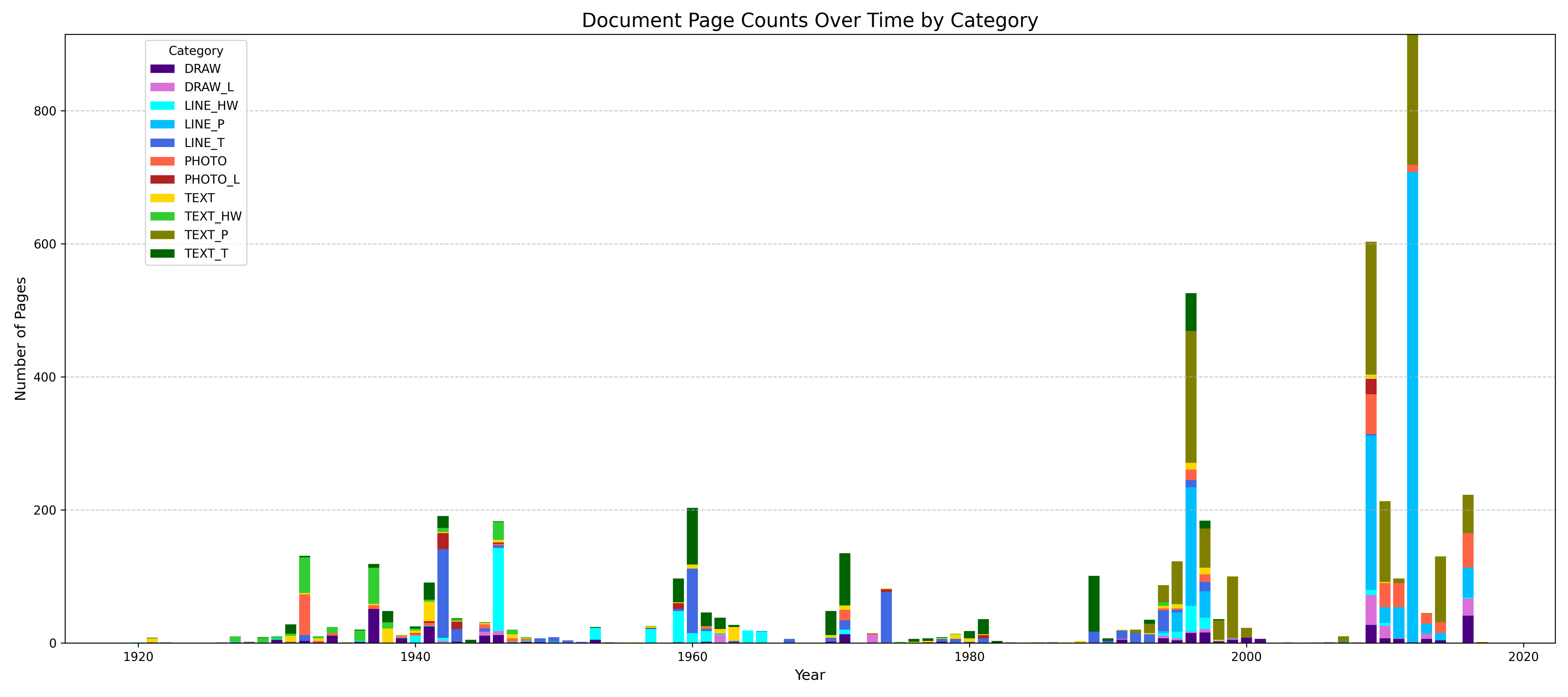}
\end{figure*}

\noindent A rigorous performance comparison was conducted across all evaluated architectures, measured by Top-1 accuracy on the 5,449-page held-out test set. The temporal distribution of this test subset is shown in Fig.~\ref{fig:label_time_test}, confirming it covers the full chronological span of the archive. The results are presented in Table~\ref{tab:model_comparison}.

\begin{figure*}[htbp]
\centering
\includegraphics[width=\textwidth]{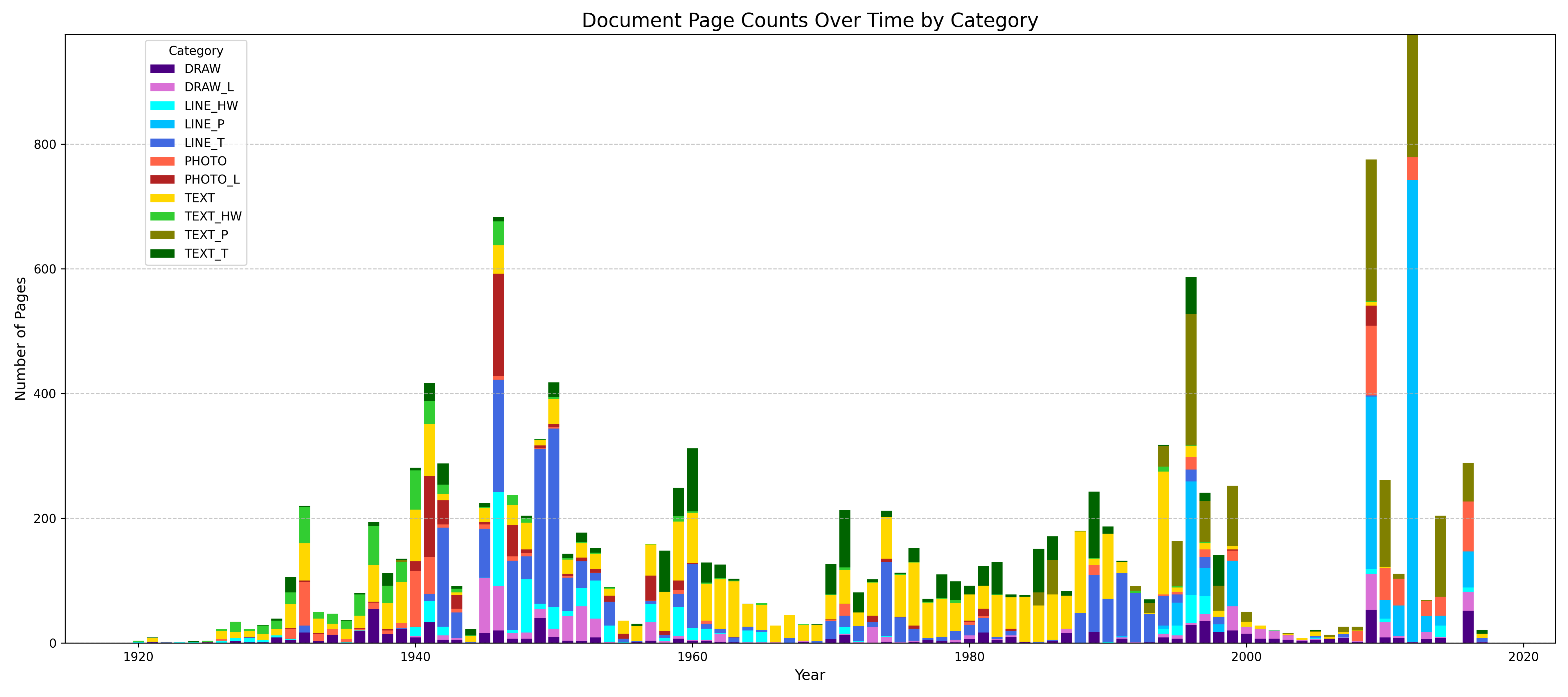}
\caption{Category distribution in the expanded test dataset (all non-training samples from the first-fold split, random seed 420; 14{,}162 pages)}
\label{fig1:time-f1}
\end{figure*}

\begin{table*}[htbp]
\centering
\caption{Comparison of Top-1 accuracy and model complexity for all evaluated architectures on the 5,449-page test set. \textit{Italic}: most parameter-efficient per model family; \textbf{bold}: best-performing (above efficiency trendline) per family. CLIP results are from the \textit{mid} description set for fine-tuned and averaged text features for zero-shot. Per-fold standard deviations and full confusion matrices are reported in~\cite{lutsai2026pageimageclassification}}
\label{tab:model_comparison}
\begin{tabular}{@{}lccc@{}}
\toprule
\textbf{Model} & \textbf{Parameters (M)} & \textbf{Input Size} & \textbf{Accuracy Top-1 (\%)} \\ \midrule
\textit{EfficientNet-v2-S} & \textit{48.2} & \textit{300$\times$300} & \textit{97.87} \\
\textbf{EfficientNet-v2-M} & \textbf{54.1} & \textbf{384$\times$384} & \textbf{98.83} \\
EfficientNet-v2-L & 118.5 & 384$\times$384 & 98.62 \\
\midrule
\textit{RegNetY-12GF} & \textit{51.8} & \textit{224$\times$224} & \textit{98.29} \\
\textbf{RegNetY-16GF} & \textbf{83.6} & \textbf{224$\times$224} & \textbf{99.16} \\
RegNetY-64GF & 281.4 & 384$\times$384 & 98.79 \\
\midrule
\textit{dit-base-rvlcdip} & \textit{86} & \textit{224$\times$224} & \textit{98.26} \\
dit-large & 304 & 224$\times$224 & 97.91 \\
\textbf{dit-large-rvlcdip} & \textbf{304} & \textbf{224$\times$224} & \textbf{98.53} \\
\midrule
\textit{vit-base-patch16} & \textit{86.6} & \textit{224$\times$224} & \textit{98.37} \\
vit-base-patch16 & 86.9 & 384$\times$384 & 98.12 \\
\textbf{vit-large-patch16} & \textbf{304.7} & \textbf{384$\times$384} & \textbf{99.12} \\
\midrule
\textbf{CLIP-ViT-B/16} & \textbf{150} & \textbf{224$\times$224} & \textbf{99.14} \\
CLIP-ViT-B/32 & 151 & 224$\times$224 & 98.99 \\
CLIP-ViT-L/14 & 428 & 224$\times$224 & 98.97 \\
CLIP-ViT-L/14 & 428 & 336$\times$336 & 98.97 \\\bottomrule
\end{tabular}
\end{table*}

Table~\ref{tab:model_comparison} demonstrates the following key findings:

\begin{itemize}
  \item \textbf{RegNetY CNN:} RegNetY-16GF achieves the highest overall accuracy ($\approx$99.16\%) with a moderate parameter count (83.6~M) at $224\times224$ resolution, outperforming the much larger 64GF variant (281.4~M). RegNetY-16GF is therefore the most efficient choice for production deployment. It also handles the dataset's hardest distinction (\texttt{TEXT} vs.\ \texttt{TEXT\_T}) best, reaching 94\% accuracy on \texttt{TEXT} with only 5.1\% confusion.
  \item \textbf{EfficientNet CNN:} EfficientNetV2-M achieves $\approx$98.83\% with only 54.1~M parameters at $384\times384$. The smaller EfficientNetV2-S (48.2~M) remains a practical option for lower-resource machines. EfficientNetV2-L underperforms its smaller sibling, confirming that capacity scaling is not monotone on this task.
  \item \textbf{Vision Transformers:} ViT-large reaches $\approx$99.12\% with 304.7~M parameters, nearly matching RegNetY-16GF at substantially higher computational cost. ViT-base at $224\times224$ achieves 98.37\% with only 86.6~M parameters.
  \item \textbf{Document Image Transformers:} DiT variants achieve 98--99\% Top-1 accuracy. Despite pre-training on document images, they do not surpass the best \acp{cnn} or ViT-large at comparable parameter budgets.
  \item \textbf{CLIP Models:} In zero-shot mode, \ac{clip}'s performance is inadequate for this domain ($<50\%$; confusion matrices in~\cite{lutsai2026pageimageclassification}). After 7~epochs of fine-tuning, the best \ac{clip} configuration (ViT-B/16 with \textit{mid} descriptions) achieves 99.14\% — competitive with RegNetY-16GF. However, see Sect.~\ref{subsec43} for an important caveat regarding \ac{clip}'s real-world deployment behavior.
\end{itemize}

Given the offline, batch-processing nature of archival workflows, the modest additional inference time for transformer-based models is acceptable. Balancing accuracy, model size, and consistency on unlabeled data, we selected \textbf{RegNetY-16GF (224)} as the primary deployment model, with \textbf{ViT-large-patch16 (384)} as a higher-resource alternative. CLIP-ViT-B/16 was released as a supplementary model for research use cases.

\paragraph{On uncertainty and comparability.} The accuracies in Table~\ref{tab:model_comparison} for image-only models are reported on the single, fixed 5,449-page test set, which is disjoint from the training subset of every cross-validation fold; the deployed weights are the element-wise average of the five per-fold checkpoints, and the per-fold accuracies underlying each averaged model, together with their spread, are tabulated in~\cite{lutsai2026pageimageclassification}. Per-model Top-1 confusion matrices for every evaluated architecture, grouped by family, are provided as Online Resource~5, with full-resolution versions in~\cite{lutsai2026pageimageclassification}. The differences among the top image-only models (RegNetY-16GF, ViT-large, EfficientNetV2-M) are small and fall within the per-fold variation, so we do not claim a statistically significant ranking among them; our deployment choice rests on the size--accuracy trade-off and, decisively, on the unlabeled-collection consistency analysis in Sect.~\ref{subsec43}. \ac{clip}, by contrast, was fine-tuned on a single fold owing to its substantially higher training cost, so no cross-fold variance is available for it. Its 99.14\% should therefore be read as a point estimate on one fold rather than a fold-averaged figure directly comparable to the image-only numbers; accordingly, our assessment of \ac{clip} relies on its deployment behaviour (Sect.~\ref{subsec43}) rather than on this test-set score.

\begin{figure*}[htbp]\centering
\includegraphics[width=130mm]{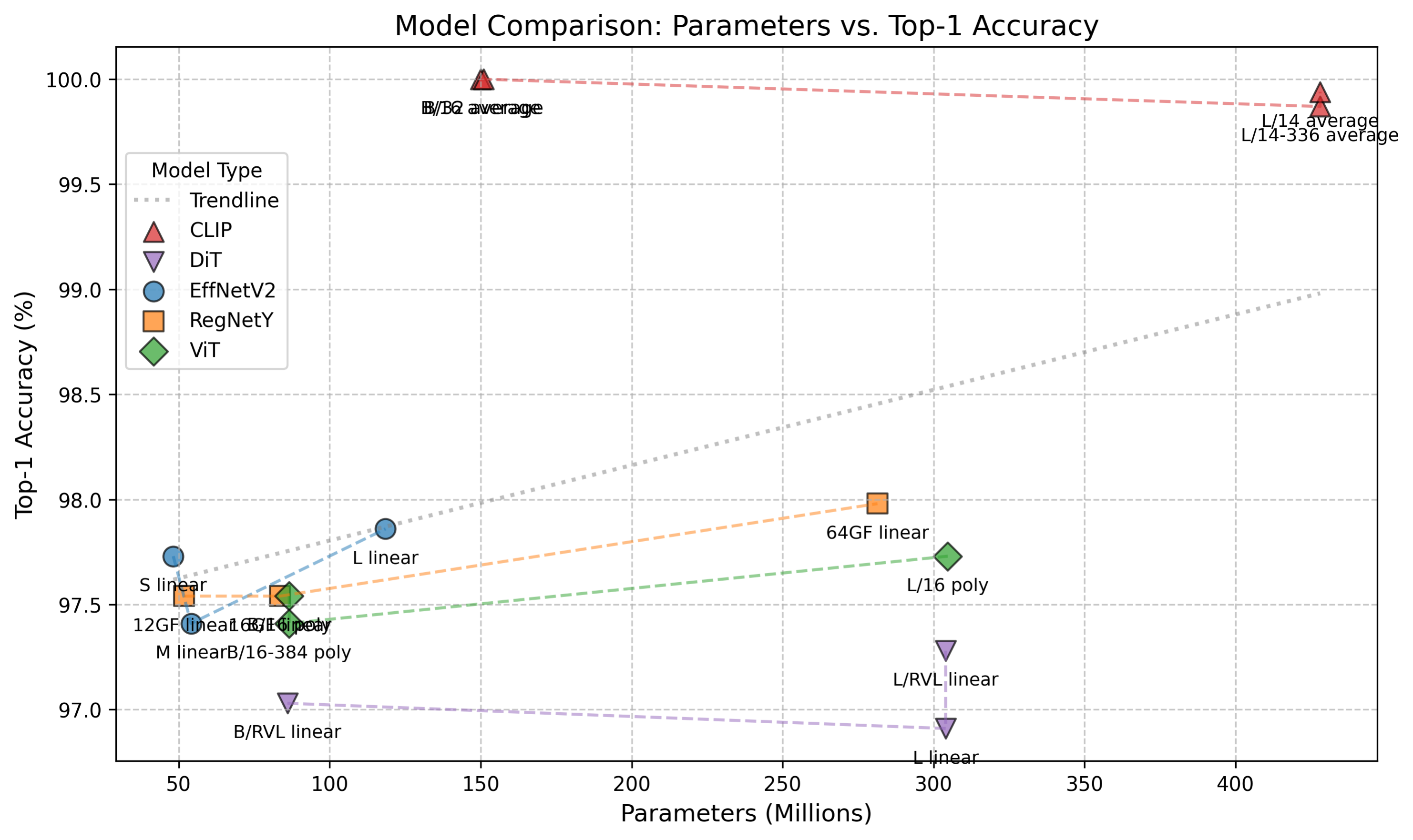}
\caption{Accuracy vs.\ Parameter Count across evaluated models. Models above the trendline deliver superior efficiency. RegNetY-16GF and CLIP ViT-B/16 both sit above the trendline, while larger CLIP-L variants fall below it}
\label{fig:acc-trend}
\end{figure*}

\subsection{Prediction Agreement on the Unlabeled Collection} \label{subsec43}

\noindent The accuracy comparisons in Sect.~\ref{subsec52} are based on the 5,449-page annotated test set. To assess behavior on real-world archival data, we additionally applied the best image-only and CLIP-based models to the full unlabeled collection of 649,508 pages and measured pairwise agreement between their predictions.

\begin{figure*}[htbp]
\centering
\includegraphics[width=130mm]{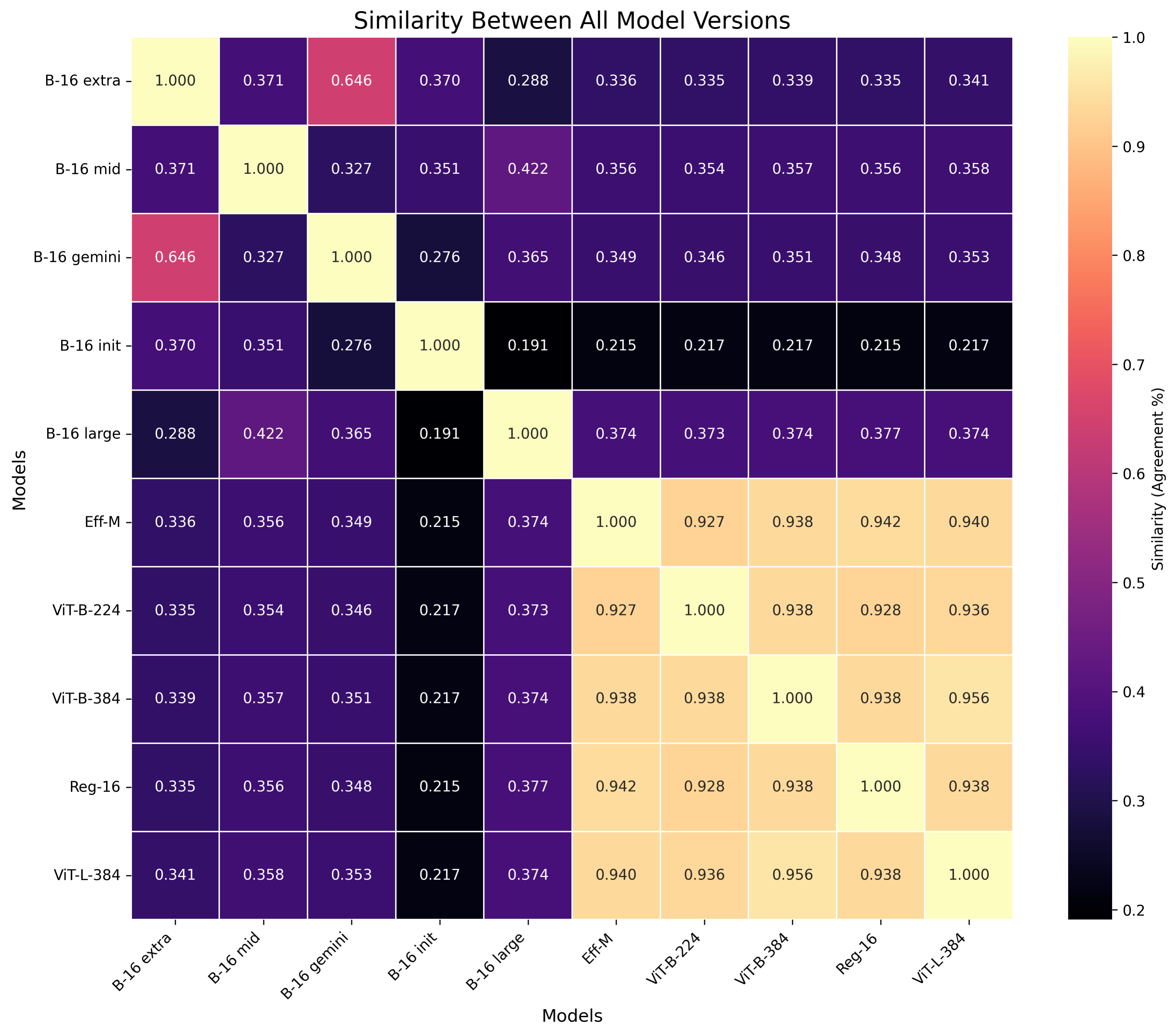}
\caption{Pairwise prediction agreement between CLIP variants (first five rows/columns: B-16 with five description sets) and image-only models (last five rows/columns) applied to the full unlabeled collection of 649,508 pages. Image-only models share over 90\% agreement; CLIP models agree with image-only models at under 65\% and show low inter-CLIP consistency}
\label{fig:similarity}
\end{figure*}

As shown in Fig.~\ref{fig:similarity}, image-only models (EfficientNetV2-M, ViT-B/224, ViT-B/384, RegNetY-16GF, ViT-large/384) share over 90\% agreement in their per-page predictions. In contrast, \ac{clip} variants agree with any image-only model at under 65\%, and even disagree substantially among themselves (inter-CLIP agreement as low as 19\% for different description sets).

The data providers confirmed that the low agreement of \ac{clip} predictions makes automated archival labeling harder to trust: \ac{clip} variants failed to assign a consistent category to over 80\% of the unlabeled pages. One likely explanation is that \ac{clip} overfits to annotated pages, inflating accuracy on the held-out test set while generalizing poorly to the full diversity of the collection. Another factor is that \ac{clip}'s multimodal text prompts may be too ambiguous for categories such as \texttt{DRAW}, which merges maps, illustrations, and schematics under a single label.

Image-only models assign consistent labels across the entire archive and produce a plausible distribution aligning with archivists' expectations. The dominant errors are concentrated in two well-understood confusions: \texttt{TEXT} vs.\ \texttt{TEXT\_T} (mixed vs.\ pure typed text) and \texttt{DRAW} vs.\ \texttt{DRAW\_L}. RegNetY-16GF performs best on both distinctions, reaching 94\% accuracy on \texttt{TEXT} and 99\% on \texttt{DRAW}.

We therefore released fine-tuned \ac{clip} variants mainly for research and illustrative purposes. For operational archival management, image-only architectures — and RegNetY-16GF in particular — are the recommended choice.

\section{Conclusion}\label{sec5}

\noindent This work addressed the critical need for automated page-image classification in heterogeneous historical archives. After establishing the unique challenges of our dataset — visual defects, skew, and mixed content types (Sect.~\ref{subsec31}) — we demonstrated the shortcomings of off-the-shelf \ac{dla} tools and a lightweight \ac{rfc} baseline, which achieved only $\sim$75\% accuracy (Sect.~\ref{subsubsec321}). This aligns with broader surveys noting the limitations of hand-crafted features~\cite{liu2021document}.

Fine-tuning modern \acp{cnn} (RegNetY and EfficientNetV2 families) markedly improved performance to $\approx$97.9--99.2\%, with RegNetY-16GF achieving the top score of 99.16\%. Fine-tuned ViT and DiT models consistently matched \acp{cnn}, with ViT-large reaching $\approx$99.12\% Top-1 accuracy (Table~\ref{tab:model_comparison}).

\ac{clip} models, once fine-tuned for 7~epochs, also achieved competitive test-set accuracy ($\approx$99.14\% for the best configuration). However, a subsequent analysis on 649,508 unlabeled pages revealed that \ac{clip} predictions diverge substantially from image-only models and show low inter-model consistency (Sect.~\ref{subsec43}), making them less suitable for large-scale archival sorting. Based on this comprehensive evaluation, \textbf{RegNetY-16GF (224)} was selected as the primary deployment model, with ViT-large-patch16 (384) as a higher-resource alternative.

\textbf{Contributions:}
\begin{itemize}
    \item Analyzed archival page-image characteristics and the failure modes of existing \ac{dla} tools.
    \item Established a resource-efficient \ac{rfc} baseline creating a $\sim$75\% accuracy benchmark on the demo category sets.
    \item Built and released an annotated dataset of over 48,000 historical page images (Dataset~3), refined through four successive annotation stages with domain-expert review~\cite{Lutsai_ATRIUM_dataset_2025}.
    \item Identified RegNetY-16GF as the primary deployment model by fine-tuning and comparing \acp{cnn}, Transformers, and multimodal architectures across five cross-validation folds.
    \item Demonstrated that image-only models produce substantially more consistent predictions on unlabeled archival data than fine-tuned \ac{clip} variants — a finding with direct implications for deployment trust.
    \item Identified EfficientNetV2-M as the best accuracy/parameter trade-off among \acp{cnn} and ViT-large as the strongest Transformer alternative; all models above the accuracy--model-size efficiency trendline are released at \href{https://huggingface.co/ufal/vit-historical-page}{ufal/vit-historical-page}~\cite{Lutsai_ATRIUM_s_page_classifier_2025}.
    \item Released fine-tuned \ac{clip} model variations at \href{https://huggingface.co/ufal/clip-historical-page}{ufal/clip-historical-page} for research purposes.
    \item Provided deployment guidance for large-scale, on-premises archival processing, including a configurable pipeline and an extensible category taxonomy.
    \item Published the dataset~\cite{Lutsai_ATRIUM_dataset_2025} and software solution~\cite{Lutsai_ATRIUM_s_page_classifier_2025} in accordance with \href{https://www.go-fair.org/fair-principles/}{FAIR} principles.
\end{itemize}

\textbf{Future Work:}
\begin{itemize}
\item Explore further architectural optimizations for improved efficiency and performance.
\item Integrate the classifier into broader digital archive management systems, noting its compatibility beyond archaeological archives.
\item Expand the category taxonomy to include additional document types (e.g., Table~\ref{tab:24-label}).
\item Incorporate larger datasets to enhance model generalization — work is already in progress at \ac{arup}.
\item Develop finer-grained \ac{clip} category descriptions (e.g.\ splitting \texttt{DRAW} into maps, schematics, and freehand sketches) to reduce multimodal model disagreement on unlabeled data.
\item Facilitate later fine-tuning stages using predictions refined by users, enhancing system adaptability over time.
\item Conduct a controlled dual-annotation study on a held-out subset to report a formal inter-annotator agreement ($\kappa$) coefficient.
\end{itemize}

By delivering near-perfect classification accuracy, consistent real-world behavior, and a practical deployment framework, this research paves the way for significantly reduced manual effort in historical document workflows, supporting digital humanities research and mass digitization initiatives~\cite{nikolaidou2022survey,liu2021document}.

\FloatBarrier
\backmatter

\section*{Statements and Declarations}\label{sec7}
\addcontentsline{toc}{section}{Statements and Declarations}

\textbf{Funding.} The conducted research received funding from the European Commission HORIZON Research and Innovation Actions under grant agreement GAP-101132163 — \href{https://atrium-research.eu/}{ATRIUM} — HORIZON-INFRA-2023-SERV-01-02 — \textbf{A}dvancing Fron\textbf{T}ier \textbf{R}esearch \textbf{I}n the Arts and h\textbf{UM}anities. Computing resources were provided by the \ac{ufal}, Charles University, via access to their \ac{cpu} and \ac{gpu} cluster nodes.

\textbf{Competing Interests.} The authors declare no competing interests.

\textbf{Author Contributions.} Kateryna Lutsai: conceptualization, data curation, software, experiments, formal analysis, writing (original draft). Pavel Straňák: supervision, project administration, review and editing. David Novák: project management, data provision, domain validation. Dana Křivánková: domain expertise, annotation review, validation, review and editing. All authors read and approved the final manuscript.

\textbf{Data Availability.} The annotated dataset is publicly available at the LINDAT repository~\cite{Lutsai_ATRIUM_dataset_2025}, containing more than 35~GB of source pages complemented with an annotation table of more than 48,000 images. The software is publicly available under the MIT license at~\cite{Lutsai_ATRIUM_s_page_classifier_2025}. Fine-tuned models are hosted on HuggingFace at \href{https://huggingface.co/ufal/clip-historical-page}{ufal/clip-historical-page} and \href{https://huggingface.co/ufal/vit-historical-page}{ufal/vit-historical-page}.

\section*{Acknowledgements}\label{sec_ack}
\addcontentsline{toc}{section}{Acknowledgements}

\noindent The project was managed by Mgr.\ David Novák, Ph.D.\ (\ac{arup}), whose coordination was instrumental throughout. The thesis on which this work is based was mentored by doc.\ RNDr.\ Pavel Pecina, Ph.D.\ (\ac{ufal}). The authors also thank the archivists of the Institutes of Archaeology in Prague and Brno for providing the source collection and for their feedback during annotation and validation.

\section*{Supplementary Information}\label{sec_si}
\addcontentsline{toc}{section}{Supplementary Information}

The following Online Resources are submitted as supplementary files accompanying this article:

\textbf{Online Resource~1} (\texttt{ESM\_1.pdf}): Visual defect examples from the archive. Includes representative samples of grayish backgrounds, page skew, bleed-through, water damage, corner holes, edge holes, stamps, fat-journal curl, manual corrections, and mixed paper textures referenced throughout Sect.~\ref{subsubsec211}.

\textbf{Online Resource~2} (\texttt{ESM\_2.pdf}): DeepDoctection (\ac{dd}) parsing attempts. Shows correct detections as well as failure modes on drawings, maps, photographs, tables, plain text, and newspaper layouts referenced in Sect.~\ref{subsec31}.

\textbf{Online Resource~3} (\texttt{ESM\_3.pdf}): Category examples. Provides multiple representative scan examples (portrait and landscape orientation) for each of the 11 classification categories (\texttt{DRAW}, \texttt{DRAW\_L}, \texttt{LINE\_HW}, \texttt{LINE\_P}, \texttt{LINE\_T}, \texttt{PHOTO}, \texttt{PHOTO\_L}, \texttt{TEXT}, \texttt{TEXT\_HW}, \texttt{TEXT\_P}, \texttt{TEXT\_T}) referenced in Sect.~\ref{subsubsec311} and Tables~\ref{tab:categories_initial}--\ref{tab:categories_used}.

\textbf{Online Resource~4} (\texttt{ESM\_4.pdf}): Dataset temporal distributions and cross-validation split proportions. Includes (a) stacked annual page counts per category across Dataset~0, Dataset~2, and Dataset~3 annotation versions (showing the characteristic 1990s gap), and (b) category proportions in train, development, and test subsets across all five cross-validation folds plotted on a document-creation-date timescale.

\textbf{Online Resource~5} (\texttt{ESM\_5.pdf}): Per-model Top-1 confusion matrices on the 5,449-page test set, grouped by architecture family: RegNetY and EfficientNetV2 (CNNs), DiT and ViT (Transformers), and fine-tuned \ac{clip} variants (averaged text features). Rows are true labels and columns predictions, normalized per row over the 11 categories, with each panel labelled by model variant and its Top-1 accuracy. Referenced in Sect.~\ref{subsec52}; full-resolution and zero-shot versions appear in~\cite{lutsai2026pageimageclassification}.

\section*{Use of AI Tools}\label{sec8}
\addcontentsline{toc}{section}{Use of AI Tools}

\noindent This manuscript was prepared with the assistance of \acp{llm}. Drafts of individual sections were generated from author-supplied bullet points and factual outlines using Gemini~2.5 and GPT-4, then reviewed and post-edited by the authors to ensure factual accuracy. Grammarly and Writefull were used for grammar and style corrections. In all cases, the authors take full accountability for the final version of the text and confirm that the edits reflect our original work. This disclosure covers the generative use of AI tools; AI-assisted copy editing per se does not require declaration per Springer Nature guidelines.

\bibliography{bibliography}

\FloatBarrier
\begin{appendices}

\renewcommand{\thefigure}{\arabic{figure}}
\renewcommand{\thetable}{\arabic{table}}
\setcounter{figure}{17}
\setcounter{table}{7}

\section{CLIP Category Descriptions}\label{secA1}

\noindent The full suite of eight category description sets used for CLIP fine-tuning and zero-shot evaluation is reproduced in the accompanying thesis~\cite{lutsai2026pageimageclassification}. Table~\ref{tab:summary_classification} summarizes all sets; Tables~\ref{tab:init_classification} and~\ref{tab:short_classification} reproduce the two most referenced sets (\textit{init} and \textit{mid}) for convenience.

\begin{table*}[htbp]
\centering
\caption{Summary of CLIP document classification description sets. ``Rev.'' denotes the revision fine-tuned to a specific label set of text features}
\label{tab:summary_classification}
\begin{tabular}{p{2cm}p{10cm}}
\toprule
\textbf{Table Label} & \textbf{Summary} \\
\midrule
init \ref{tab:init_classification} &
Provides the full ``initial'' set of classification categories with detailed distinctions between drawings, photos, and text, and further separates content by handwritten (HW), printed (P), and typed (T) forms both inside and outside tables or forms. \\[1ex]

mid \ref{tab:short_classification} &
Offers a more concise restatement of the initial taxonomy, trimming phrasing while preserving the same category distinctions, emphasizing brevity for quick reference. \\[1ex]

min &
Distills the taxonomy to its bare essentials, reducing each description to the minimal wording needed to convey whether an element is a drawing, table, photo, or text and its modality. \\[1ex]

gpt &
Reframes the classification around page-level context, explicitly noting that each label applies to an entire page containing the specified content (proposed by GPT-4 Deep Research). \\[1ex]

short &
Adapts the minimal set by omitting the word ``page'' and refining descriptions to emphasize cropped or cell-level occurrences. \\[1ex]

gemini &
Expands upon the page-based version with verbose, researcher-oriented descriptions, adding more examples and elaboration per category (proposed by Gemini 2.5 Deep Research). \\[1ex]

extra &
Presents a balanced yet thorough taxonomy enriched with illustrative examples. \\[1ex]

detailed &
Delivers the most detailed annotation-driven version, augmenting each label with notes on layout, annotation styles, and use-case examples. \\
\bottomrule
\end{tabular}
\end{table*}

\begin{table*}[htbp]
\centering
\caption{Initial document classification labels (\textit{init} set)}
\label{tab:init_classification}
\begin{tabular}{>{\raggedright\arraybackslash}p{2.2cm}>{\raggedright\arraybackslash}p{10cm}}
\toprule
\textbf{Label} & \textbf{Description} \\
\midrule
DRAW & drawings, maps, paintings, schematics, graphics with labels \\
DRAW\_L & drawings, maps, paintings, schematics, graphics with a table legend, inside a table or form \\
LINE\_HW & handwritten text lines inside a table or form \\
LINE\_P & printed text lines inside a table or form \\
LINE\_T & typed text lines inside a table or form \\
PHOTO & photos or cutouts from photos with labels \\
PHOTO\_L & photos with a table caption, inside a table or form \\
TEXT & mixed printed and handwritten texts \\
TEXT\_HW & handwritten text page \\
TEXT\_P & printed text page \\
TEXT\_T & typed document page \\
\bottomrule
\end{tabular}
\end{table*}

\begin{table*}[htbp]
\centering
\caption{Short document classification labels (\textit{mid} set) — the best-performing CLIP configuration}
\label{tab:short_classification}
\begin{tabular}{>{\raggedright\arraybackslash}p{2.2cm}>{\raggedright\arraybackslash}p{9.8cm}}
\toprule
\textbf{Label} & \textbf{Description} \\
\midrule
DRAW & a drawings or a map or a diagram \\
DRAW\_L & a table and drawings or a map or a diagram \\
LINE\_HW & a table or a filled form of handwritten texts \\
LINE\_P & a table or a filled form of printed texts \\
LINE\_T & a table or a filled form of typed texts \\
PHOTO & photos or photo cutouts \\
PHOTO\_L & a table and photos or photo cutouts \\
TEXT & printed and handwritten text styles on a page \\
TEXT\_HW & a handwritten plain text page or handwritten text paragraphs \\
TEXT\_P & a printed plain text page or printed text paragraphs \\
TEXT\_T & a typed plain text document page or typed text paragraphs \\
\bottomrule
\end{tabular}
\end{table*}

\section{System Overview}\label{secC1}

\noindent The system's primary entry point is the \texttt{run.py} script, which provides a comprehensive \ac{cli} for granular control over the system's operations. The available flags are organized by user group, as shown in Table~\ref{tab:cli-options}.

\begin{table*}[htbp]
\centering
\caption{CLI options for \texttt{run.py}}
\label{tab:cli-options}
\begin{tabular}{lllp{7cm}}
\toprule
\textbf{Group} & \textbf{Flag} & \textbf{Alias} & \textbf{Description} \\
\midrule
\multicolumn{4}{l}{\textbf{USER}} \\
& \texttt{--help} & \texttt{-h} & Displays available command options \\
& \texttt{--hf} & & Interfaces with Hugging Face for model retrieval/publishing \\
& \texttt{--revision} & \texttt{-rev} & Specifies the model version \\
& \texttt{--model} & \texttt{-m} & Sets the base model architecture \\
& \texttt{--file} & \texttt{-f} & Processes a single image file \\
& \texttt{--directory} & \texttt{-d} & Explicitly specifies an input directory \\
& \texttt{--dir} & & Uses the default directory from the configuration file \\
& \texttt{--topn} & \texttt{-tn} & Defines the number of top-N guesses to return \\
& \texttt{--raw} & & Generates raw probability distributions for predictions \\
& \texttt{--batch\_size} & & Sets the batch size for training, evaluation, and inference \\
\midrule
\multicolumn{4}{l}{\textbf{CLIP-specific}} \\
& \texttt{--avg} & & Uses an average of text features from all category descriptions \\
& \texttt{--model\_path} & & Specifies the filename of a local \texttt{.pt} model checkpoint \\
\midrule
\multicolumn{4}{l}{\textbf{DEV}} \\
& \texttt{--train} & & Initiates a model training session \\
& \texttt{--eval} & & Performs model evaluation on a dataset \\
& \texttt{--lr} & & Sets the learning rate for training \\
& \texttt{--epochs} & & Sets the number of epochs for training \\
& \texttt{--max\_categ} & \texttt{-mc} & Sets the maximum samples per category for a training subset \\
& \texttt{--max\_categ\_eval} & \texttt{-mce} & Sets the maximum limit of samples per category for evaluation \\
& \texttt{--model\_dir} & & Specifies path to a directory of saved model checkpoints \\
& \texttt{--eval\_dir} & & Evaluates a directory of saved models \\
\bottomrule
\end{tabular}
\end{table*}

\begin{figure*}[htbp]
\centering
\includegraphics[width=0.75\textwidth]{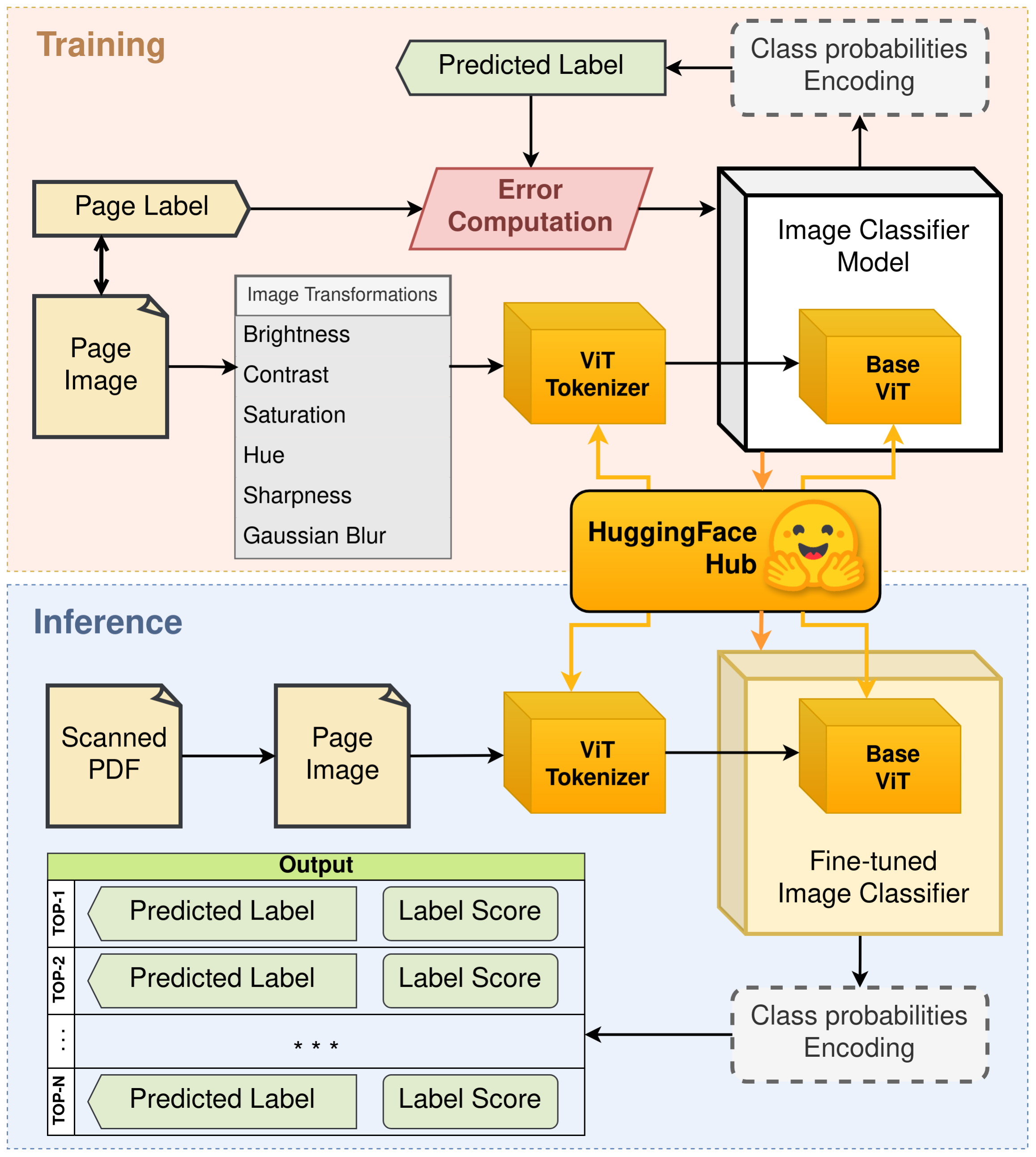}
\caption{Scheme of the Transformers and CNNs model architecture}
\label{fig:architect_trans}
\end{figure*}

\begin{figure*}[htbp]
\centering
\includegraphics[width=0.75\textwidth]{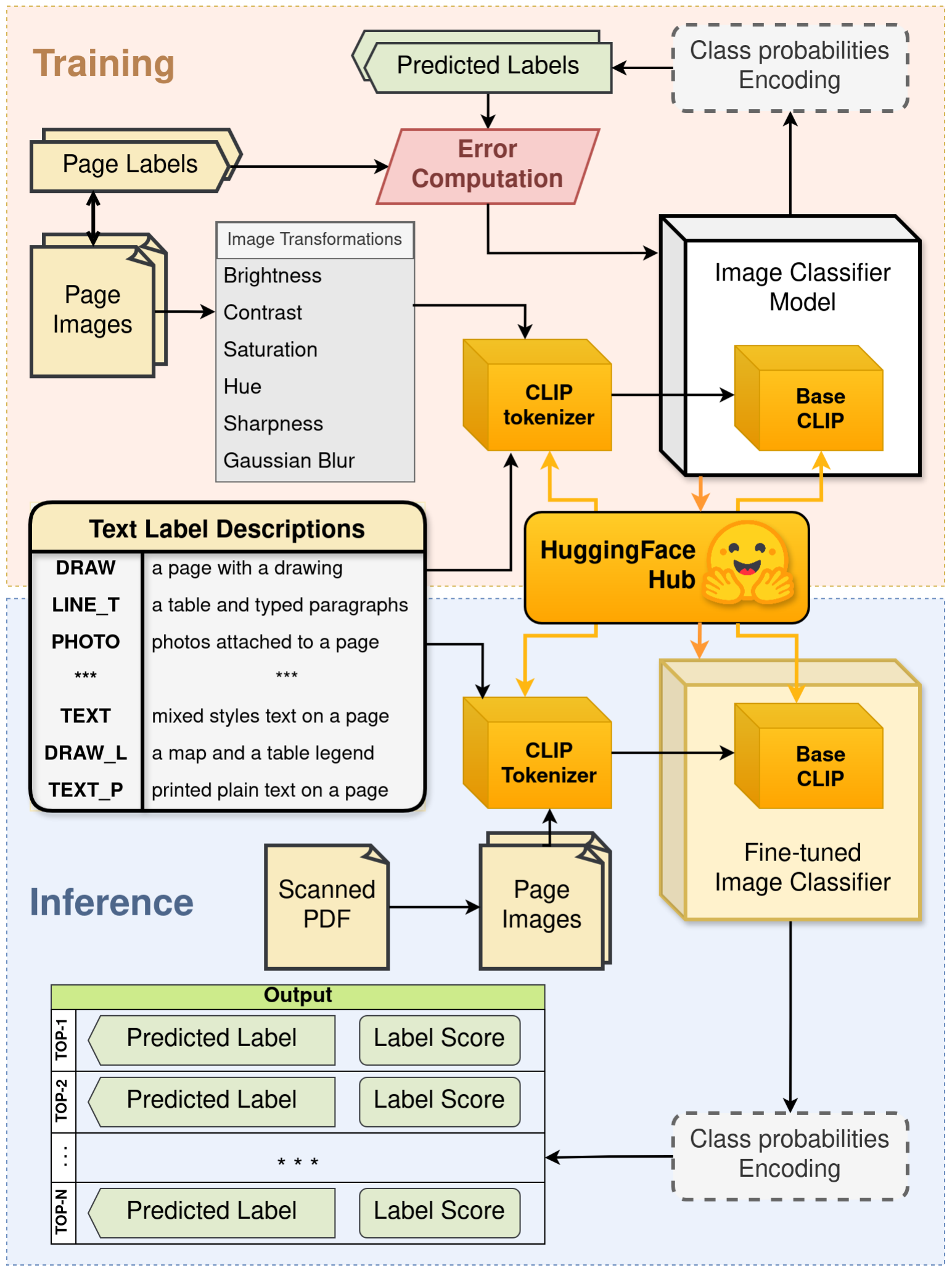}
\caption{Scheme of the CLIP model architecture}
\label{fig:architect_clip}
\end{figure*}

\begin{table*}[htbp]
\centering
\small
\setlength{\tabcolsep}{4pt}
\renewcommand{\arraystretch}{0.9}
\begin{tabular}{r|cccc|}
 & \multicolumn{1}{c|}{Printed} & \multicolumn{1}{c|}{Typewritten} & \multicolumn{1}{c|}{HandWritten} & Mixed \\ \hline
Photos & \multicolumn{4}{c|}{\textbf{PHOTO}} \\ \hline
Drawings etc. & \multicolumn{4}{c|}{\textbf{DRAW}} \\ \hline
Photo in table & \multicolumn{4}{c|}{\textbf{PHOTO\_L}} \\ \hline
Drawing in table & \multicolumn{4}{c|}{\textbf{DRAW\_L}} \\ \hline
Tables \& Forms & \multicolumn{1}{c|}{\textbf{LINE\_P}} & \multicolumn{1}{c|}{\textbf{LINE\_T}} & \multicolumn{2}{c|}{\textbf{LINE\_HW}} \\ \hline
Plain texts & \multicolumn{1}{c|}{\textbf{TEXT\_P}} & \multicolumn{1}{c|}{\textbf{TEXT\_T}} & \multicolumn{1}{c|}{\textbf{TEXT\_HW}} & \textbf{TEXT}
\end{tabular}
\caption{Defined labels coverage of the data features variability}
\label{tab:11-label}
\end{table*}

\begin{table*}[htbp]
\centering
\small
\setlength{\tabcolsep}{4pt}
\renewcommand{\arraystretch}{0.9}
\begin{tabular}{p{1.6cm}|c|c|c|c|}
 & Printed & Typewritten & Handwritten & Mixed \\ \hline
Photos & \textbf{P\_P} & \textbf{P\_T} & \textbf{P\_HW} & \textbf{PHOTO} \\ \hline
Drawings & \textbf{D\_P} & \textbf{D\_T} & \textbf{D\_HW} & \textbf{DRAW} \\ \hline
Photos in table & \multicolumn{1}{l|}{\textbf{P\_L\_P}} & \multicolumn{1}{l|}{\textbf{P\_L\_T}} & \multicolumn{1}{l|}{\textbf{P\_L\_HW}} & \multicolumn{1}{l|}{\textbf{PHOTO\_L}} \\ \hline
Drawings in table & \textbf{D\_L\_P} & \textbf{D\_L\_T} & \textbf{D\_L\_HW} & \multicolumn{1}{l|}{\textbf{DRAW\_L}} \\ \hline
Tables \& Forms & \textbf{LINE\_P} & \textbf{LINE\_T} & \textbf{LINE\_HW} & \textbf{LINE} \\ \hline
Plain texts & \textbf{TEXT\_P} & \textbf{TEXT\_T} & \textbf{TEXT\_HW} & \textbf{TEXT}
\end{tabular}
\caption{Expanded labels coverage of the data features variability}
\label{tab:24-label}
\end{table*}

\end{appendices}

\end{document}